%% file: neurips_2026.tex
\documentclass{article}

\usepackage[main, preprint]{neurips_2026}

\usepackage[utf8]{inputenc}
\usepackage[T1]{fontenc}
\usepackage{hyperref}
\usepackage{url}
\newcommand{\codeurl}{\url{https://github.com/Ming-321/PISG}}
\usepackage{booktabs}
\usepackage{multirow}
\usepackage{amsfonts}
\usepackage{nicefrac}
\usepackage{microtype}
\usepackage{xcolor}
\usepackage{graphicx}
\usepackage{amsmath}
\usepackage[hypcap=false]{caption}
\usepackage{wrapfig}
\usepackage{enumitem}
\usepackage{algorithm}
\usepackage{algorithmic}
\setlist[itemize]{leftmargin=1.5em, itemsep=2pt, parsep=0pt, topsep=2pt}
\setlist[enumerate]{leftmargin=1.8em, itemsep=2pt, parsep=0pt, topsep=2pt}

\title{PISG: Constraint-Aligned Signal Amplification for Diffusion-Based Combinatorial Optimization}

\author{%
  Yuming Zhang \quad Xianchen Zhou \quad Hongxia Wang\thanks{Corresponding author.} \\
  College of Science, National University of Defense Technology, Changsha, China \\
  \texttt{\{zhangyuming, zhouxianchen13, wanghongxia\}@nudt.edu.cn}
}

\begin{document}

\maketitle

\begin{abstract}
\input{sec/abstract}
\end{abstract}

\input{sec/introduction}
\input{sec/related_work}
\input{sec/preliminaries}
\input{sec/method}
\input{sec/signal_characterization}
\input{sec/experiments}
\input{sec/conclusion}

\begin{ack}
We would like to express our gratitude to Prof.\ Guangquan Cheng and Changjun Fan for their insightful comments.
The work was supported by the National Natural Science Foundation of China (No.~12401677 and 12471401), and Youth Science Fund of National University of Defense Technology (No.~ZK23-07).
The authors declare no competing interests.
\end{ack}

\bibliographystyle{abbrvnat}
\bibliography{references}

\appendix

\paragraph{Appendix overview.}
The appendix is organized as follows.
Appendix~\ref{app:method-details} provides implementation details for reproducibility (Section~\ref{sec:method}) and applies PISG to the Transformer-based GenSCO solver (Appendix~\ref{app:gensco}).
Appendix~\ref{app:full-results} documents full results with standard deviations and statistical tests, the deployment pipeline with objective rewrite and 2-opt, and compute cost with compute-matched comparisons (Sections~\ref{sec:exp:main}--\ref{sec:exp:efficiency}).
Appendix~\ref{app:config-selection} reports configuration selection, sensitivity, within-family transfer, and transfer under distribution shift (Section~\ref{sec:exp:sensitivity}).
Appendix~\ref{app:mechanism} presents additional mechanism evidence and the limits of the diagnostics (Section~\ref{sec:signal-characterization}).

\input{appendix/appendix}

\end{document}

%% file: sec/abstract.tex
Diffusion-based solvers have become a prominent approach to neural combinatorial optimization.
At inference time, they assign each decision variable a prediction score.
A greedy decoder ranks candidates by these scores and checks hard constraints to ensure feasibility.
Top-ranked candidates should be both promising and jointly feasible, making prediction--constraint alignment critical.
When this alignment is weak, greedy decoding skips constraint-violating candidates and falls back to lower-ranked alternatives, degrading solution quality.
Existing test-time methods improve search, sampling, or decoding, but do not directly strengthen this alignment before decoding.
We observe that trained denoisers already carry constraint-aligned structural signals.
Building on this observation, we propose Perturbed Instance-Structure Guidance (PISG), a training-free method that amplifies these signals at test time.
At each denoising step, PISG constructs a structure-degraded prediction and extrapolates the unperturbed prediction away from it.
The resulting guided prediction improves candidate rankings before constrained decoding.
Signal analysis shows that this guidance concentrates on constraint-critical variables rather than uniformly rescaling scores.
Under greedy decoding and without downstream refinement, PISG improves all 16 TSP solver--scale pairs across DIFUSCO, T2T, FastT2T, and StruDiCO, and all four MIS-ER solvers, with MIS-ER solved-cost gains up to $+4.8\%$.
The gains persist on most large-scale settings after 2-opt, and a preliminary study extends PISG to a Transformer-based solver.
These results identify prediction--constraint alignment as an effective test-time enhancement direction for diffusion-based combinatorial optimization.
Code is available at \codeurl.

%% file: sec/introduction.tex
\section{Introduction}
\label{sec:introduction}

Diffusion-based solvers have become a prominent approach to neural combinatorial optimization~\citep{sun2023difusco, li2023t2t, li2024fastt2t, wang2025strudico}.
At inference time, the denoiser does not directly return the final feasible solution.
Instead, it assigns each candidate a prediction score.
For the Travelling Salesman Problem (TSP), the candidates are edges that may enter the tour; for the Maximum Independent Set problem (MIS), they are vertices that may enter the independent set.
A greedy decoder considers candidates in descending score order and accepts each one that passes hard constraint checks.
Feasibility is therefore guaranteed by the decoder, but the final solution quality is still shaped by the ranking produced before decoding.
Top-ranked candidates should be both promising and jointly feasible.
This makes prediction--constraint alignment before decoding a central factor in solution quality.

\begin{figure}[t]
\centering
\includegraphics[width=0.9\linewidth]{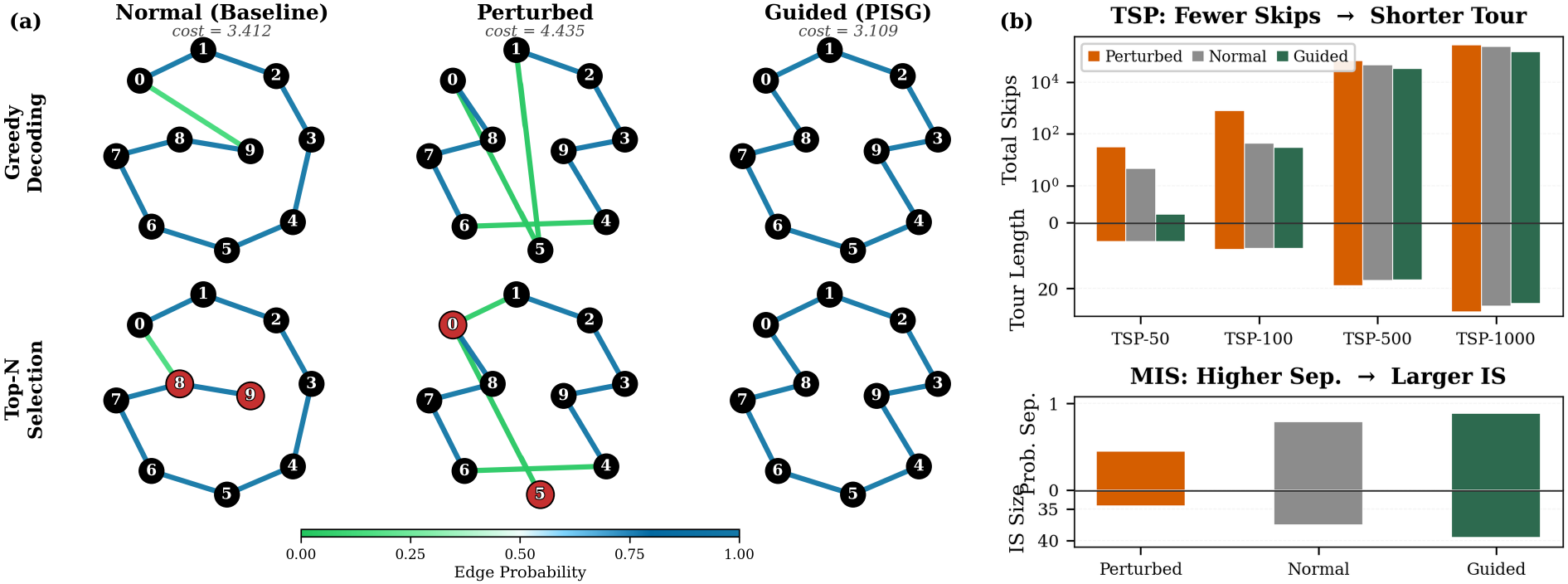}
\caption{Prediction quality governs decoder skips and final solution quality.
\textbf{(a)}~Schematic 10-node TSP (synthetic, not solver output). \emph{Top row}: greedy tours are always feasible, but lower prediction quality leads to costlier tours ($3.109 < 3.412 < 4.435$).
\emph{Bottom row}: top-$N$ selected edges. Normal already concentrates predictions on degree-compatible edges; weakening edge gating (Perturbed) visibly increases violations; Guided further reduces them to zero (no red nodes).
\textbf{(b)}~Quantitative evidence across four TSP scales and MIS: fewer decoder skips correspond to shorter tours and larger independent sets, confirming that prediction quality before decoding governs final solution quality.
See Appendix~\ref{app:raw-output-alignment}.}
\label{fig:motivation}
\end{figure}

A decoder skip is the clearest symptom of weak prediction--constraint alignment.
It occurs when a high-ranked candidate conflicts with previously accepted choices and is rejected by the feasibility check.
Figure~\ref{fig:motivation}(a) makes this concrete in a schematic TSP-10 instance.
High-ranked candidate edges are shown in blue; green edges are lower-ranked fallback alternatives, chosen only after higher-ranked edges fail constraint checks.
When feasibility checks skip a high-ranked edge, the decoder continues down the ranked list until it finds an edge that can still be inserted.
In this example, fallback edges are longer, so a local ranking error becomes a longer final tour even though feasibility is preserved.
Each skip is thus a compromise: in TSP it substitutes a costlier edge; in MIS it forgoes a node that could have enlarged the independent set.
Figure~\ref{fig:motivation}(b) quantifies this pattern: across four TSP scales, skips decrease and tours shorten monotonically as prediction quality improves (at TSP-50, skips drop from $32.4$ to $0.08$); MIS probability separation and independent-set size follow the same ordering.
These skip patterns motivate an upstream test-time target: improving prediction--constraint alignment before greedy decoding.

Several test-time strategies have improved diffusion-based CO solvers by targeting downstream or trajectory-level components.
Objective-guided rewrite refines decoded solutions~\citep{li2023t2t}; consistency acceleration reduces trajectory length~\citep{li2024fastt2t}; structured noising with objective-aware decoding reshapes the sampling process~\citep{wang2025strudico}.
None directly addresses how well the denoiser separates feasible from infeasible candidates at constraint boundaries.
We refer to this separation as \emph{constraint-boundary discriminability}: the degree to which the denoiser's predictions distinguish between candidates that should and should not be selected near hard constraints.
Higher discriminability means fewer decoder skips, making it an upstream channel that remains largely unexploited.

Through training on feasible solutions, the denoiser produces predictions that already reflect constraint structure before any decoding.
The bottom row of Figure~\ref{fig:motivation}(a) selects top-$N$ edges by predicted score alone, without any constraint enforcement.
Even so, Normal already achieves near degree-2 compliance, closely matching what greedy decoding would accept.
This suggests that trained denoisers have implicitly acquired sensitivity to constraint structure, a signal that might be amplified at test time.
We study this within the \emph{edge-gated AGNN family}---the denoisers underlying DIFUSCO, T2T, FastT2T, and StruDiCO~\citep{sun2023difusco, li2023t2t, li2024fastt2t, wang2025strudico}---which share a structure-sensitive mechanism called \emph{edge gating} (Section~\ref{sec:preliminaries:gating})~\citep{bresson2017residual, sun2023difusco}.
Edge gating is the primary pathway through which instance-specific graph structure enters the prediction.
Figure~\ref{fig:motivation}(a) shows this in a synthetic example: weakening edge gating produces predictions with more degree violations and a costlier greedy tour.
Since this sensitivity enters through edge gating, weakening it should expose the direction along which instance structure shapes predictions.

We propose \emph{Perturbed Instance-Structure Guidance (PISG)} to exploit this structural sensitivity.
It follows the perturb-and-extrapolate form of visual diffusion guidance~\citep{ahn2024pag, hong2023improving}, but changes the perturbation target to edge gating, the instance-structure interface in edge-gated AGNN denoisers.
The extrapolation rule is thus inherited; our contribution is a CO-specific intervention on the structure pathway, together with evidence that the direction it induces is constraint-aligned.
PISG replaces the learned edge gates with a weakened surrogate, forming a structure-degraded branch that shares all inputs and parameters with the normal branch.
Extrapolating the normal prediction away from the degraded one amplifies the structure-dependent component of each denoising step.
We verify in Section~\ref{sec:signal-characterization} that this direction is constraint-aligned: it concentrates at task-specific constraint boundaries rather than rescaling predictions isotropically.
PISG is training-free and uses no task objective, but it is not tuning-free: its scale and perturbation strategy are selected by a modest validation sweep for each task--distribution pair.
The two parameter-sharing branches can be batched during inference.

Our experiments examine PISG at the signal level and after greedy decoding.
Signal characterization shows that the extrapolation direction carries constraint-boundary discriminability: in TSP, it separates degree-2 boundary edges, and in MIS, it separates conflicting vertex choices (Section~\ref{sec:signal-characterization}).
Under greedy decoding without objective-guided refinement, PISG improves all 16 TSP solver--scale pairs and all four MIS-ER solvers, with MIS-ER solved-cost gains up to \(+4.8\%\).
Additional experiments show that this constraint-aligned signal is mechanistically distinct from objective-side refinement; where residual headroom remains, combining the two can yield further gains (Section~\ref{sec:exp:deployment}).
With 2-opt added, gains persist in three of four large-scale cells but become much smaller, and a preliminary transfer to a Transformer solver shows similarly condition-dependent effects.
Our contribution is threefold.
First, we identify prediction--constraint alignment as a pre-decoding bottleneck in diffusion-based combinatorial optimization and formalize its local manifestation as constraint-boundary discriminability.
Second, we propose PISG, a training-free guidance method that perturbs edge-gated instance-structure processing and extrapolates away from the resulting degraded prediction at each denoising step, before greedy decoding.
Third, we show that PISG amplifies constraint-aligned signals at TSP and MIS constraint boundaries rather than uniformly rescaling predictions; under greedy decoding without downstream refinement, this yields gains across all four edge-gated solvers.

%% file: sec/related_work.tex
\section{Related Work}
\label{sec:related}

\paragraph{Solver-level advances in diffusion-based CO.}
DIFUSCO introduced discrete diffusion over graph representations for CO \citep{sun2023difusco}.
T2T added objective-guided rewrite (\(T_g\)) that iteratively improves solutions through gradient-guided denoising \citep{li2023t2t}.
FastT2T replaced iterative diffusion with optimization consistency training, significantly shortening inference \citep{li2024fastt2t}.
StruDiCO redesigned the noise and reverse processes via variable-absorption noising and introduced gradient-free objective-aware refinement \citep{wang2025strudico}.
Each targets a different axis of the pipeline: training targets, noise processes, or inference strategy.
Yet all four build on the same edge-gated AGNN backbone \citep{bresson2017residual} and follow a common skeleton of denoising followed by greedy decoding.
Concurrent work extends this line with search-driven generation~\citep{li2025gensco}, adaptive expansion~\citep{ma2025coexpander}, masked generation and diffusion~\citep{chen2026maskco, wang2026nexco}, and energy-guided adaptation~\citep{lei2025difuada}.
All of these innovations target trajectory design, refinement policy, or search strategy, treating the denoiser's per-step predictions as given.
None explicitly targets the denoiser's constraint-boundary discriminability at the prediction level.

\paragraph{Denoiser-internal guidance.}
Image generation provides a precedent: several methods modify the denoiser's internal computation to steer sampling.
Classifier guidance steers diffusion sampling with an external gradient \citep{dhariwal2021diffusion}.
Classifier-free guidance (CFG) replaces it by contrasting conditional and unconditional predictions but requires training-time modifications \citep{ho2022cfg}.
Recent CFG extensions for discrete diffusion share this training requirement \citep{schiff2025simple, he2026guidance}.
Training-free variants go further: PAG, SAG, SEG, and CDG perturb internal components (e.g., self-attention maps or condition embeddings) to construct a degraded branch and extrapolate \citep{ahn2024pag, hong2023improving, hong2024smoothed, han2026cdg}; see also \citep{ahn2025headhunter, bansal2023universal, chung2024selfguidance, ye2024tfg}.
These training-free methods share a common design template: they construct a degraded version of the denoiser by perturbing the mechanism that selectively routes information based on input content or structure (self-attention in image U-Nets and DiTs), then extrapolate away from the degraded prediction.
In graph-based CO denoisers, learned edge gates rather than self-attention serve as the primary information-routing mechanism, selectively scaling neighbor contributions during message passing.
Whether perturbing this pathway can reveal constraint-aligned sensitivity in CO denoisers has not been tested.
We investigate this question within the edge-gated AGNN family (DIFUSCO, T2T, FastT2T, StruDiCO), where the shared gating mechanism enables controlled isolation of the denoiser-internal effect.
We additionally report a preliminary transfer to GenSCO~\citep{li2025gensco}, which routes structure through attention biases.
Our signal characterization shows that the resulting signal concentrates at constraint boundaries rather than rescaling isotropically, connecting the visual-domain template to a CO-specific, constraint-aligned finding.

%% file: sec/preliminaries.tex
\section{Preliminaries}
\label{sec:preliminaries}

\paragraph{Task definitions.}
Given a graph \(G=(V,E)\), a combinatorial optimization (CO) problem seeks a binary vector \(\mathbf{x}\in\{0,1\}^{N}\) that optimizes an objective \(f(\mathbf{x};G)\) subject to feasibility constraints \(\Omega\).
\textbf{TSP} is an edge-selecting problem on a complete graph of \(n\) Euclidean nodes.
The goal is to find a minimum-cost Hamiltonian cycle.
Every node must be incident to exactly two tour edges (the \emph{degree-2} constraint).
\textbf{MIS} is a node-selecting problem on a given graph.
The goal is to find a largest independent set.
No two adjacent nodes may be simultaneously selected (\emph{pairwise exclusion}).
These constraints define the hard feasibility checks used by the constrained decoder and the constraint boundaries analyzed in Section~\ref{sec:signal-characterization}: the degree-2 boundary in TSP and adjacency conflicts in MIS.

\paragraph{Diffusion-based CO pipeline.}
Recent diffusion and consistency solvers learn solution distributions conditioned on the instance graph~\citep{sun2023difusco, li2023t2t, li2024fastt2t, wang2025strudico}.
Inference starts from a solver-specific terminal state \(\mathbf{x}_{T_s}\) (uniformly random binary for standard discrete-diffusion solvers; see Appendix~\ref{app:full-results} for StruDiCO) and proceeds through \(T_s\) denoising steps.
At each step \(t\), a trained denoiser receives the noisy state \(\mathbf{x}_t\), the denoising index \(t\), and the instance graph \(G\), and outputs \(N\) per-element logit scores \(\mathbf{p}_t\) estimating the clean solution \(\mathbf{x}_0\).
After the final step, these scores feed the greedy constrained decoder.
Optional \(T_g\) objective-guided rewrite steps operate downstream of decoding~\citep{li2023t2t, li2024fastt2t, wang2025strudico}.

\paragraph{AGNN edge gating.}
\label{sec:preliminaries:gating}
The denoisers in DIFUSCO, T2T, FastT2T, and StruDiCO share an anisotropic GNN backbone~\citep{bresson2017residual}.
Each layer updates node features via edge-gated message passing
\begin{equation}
\label{eq:agnn}
\mathbf{h}_{i}^{(\ell+1)}
=
\mathbf{h}_{i}^{(\ell)}
+
\mathrm{ReLU}\!\left(
\mathrm{LN}\!\left(
\mathbf{U}^{(\ell)}\mathbf{h}_{i}^{(\ell)}
+
\operatorname{Aggr}_{j \in \mathcal{N}(i)}\!
\left[
\sigma\!\left(\hat{\mathbf{e}}_{ij}^{(\ell+1)}\right)
\odot
\mathbf{V}^{(\ell)}\mathbf{h}_{j}^{(\ell)}
\right]
\right)
\right),
\end{equation}
where \(\hat{\mathbf{e}}_{ij}^{(\ell+1)}\) is a pre-activation edge feature aggregating node states and edge attributes via learned projections~\citep{sun2023difusco}, \(\mathbf{U}^{(\ell)}\) and \(\mathbf{V}^{(\ell)}\) are node projection matrices, and \(\mathrm{LN}\) denotes layer normalization.
We denote the \emph{edge gate} \(\mathbf{g}_{ij}^{(\ell)} = \sigma(\hat{\mathbf{e}}_{ij}^{(\ell+1)})\); it selectively scales each neighbor's projected contribution \(\mathbf{V}^{(\ell)}\mathbf{h}_{j}^{(\ell)}\), determining which pairwise relations the denoiser emphasizes at each layer.
We collect all layer--edge gates into a map \(\Gamma=\{\mathbf{g}_{ij}^{(\ell)}\}_{\ell,(i,j)}\), the structure-sensitive internal pathway that PISG perturbs in Section~\ref{sec:method}.

%% file: sec/method.tex
\section{Method: Constraint-Aligned Signal Amplification via PISG}
\label{sec:method}

\begin{figure}[t]
\centering
\includegraphics[width=0.9\linewidth]{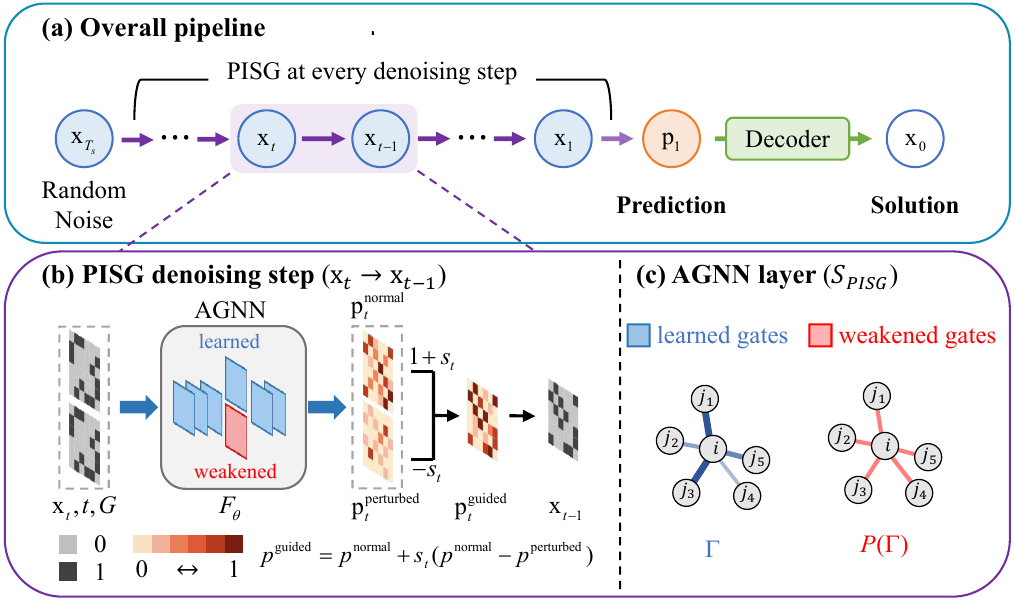}
\caption{Overview of PISG.
(a) PISG is applied at every denoising step to form guided logits for the reverse transition.
(b) At step $t$, PISG computes two parallel, parameter-sharing denoising branches: a normal branch with learned AGNN gates and a structure-degraded branch with weakened gates. Their logit discrepancy defines the PISG signal, and the guided logits extrapolate along this direction (Equation~\ref{eq:pisg}) for sampling $x_{t-1}$.
(c) In selected AGNN layers, PISG replaces heterogeneous learned edge gates with spatially uniform weakened surrogates, removing edge-wise structural sensitivity.}
\label{fig:pisg-overview}
\end{figure}

PISG is a training-free mechanism for constraint-aligned signal amplification before constrained decoding.
As shown in Figure~\ref{fig:pisg-overview}, it targets the denoiser's edge-gated instance-structure pathway and constructs a structure-degraded branch at each denoising step.
The normal-minus-perturbed discrepancy is then used to amplify the structure-dependent component of the prediction before the decoder ranks candidates.

\paragraph{Why target edge gating?}
In the AGNN backbone (Section~\ref{sec:preliminaries:gating}), the edge gate is the primary pathway through which instance-specific graph structure shapes per-step predictions.
Weakening this pathway degrades structure-sensitive message routing while keeping the input graph, noisy state, model parameters, and decoder unchanged.
The resulting difference is designed to isolate the denoiser's structure-dependent sensitivity.
If this sensitivity is aligned with hard constraints, the normal-minus-perturbed direction should improve constraint-boundary discriminability, which we test in Section~\ref{sec:signal-characterization}.

\subsection{Two-Branch Construction and Guidance Signal}
\label{sec:method:guidance}

Let $F_\theta(\mathbf{x}_t, t, G;\, \Gamma)$ denote a denoiser forward pass that receives the noisy state $\mathbf{x}_t$, denoising step $t$, instance graph $G$, and the internal edge-gate map $\Gamma=\{\mathbf{g}_{ij}^{(\ell)}\}$.
The notation $F_\theta(\cdot;\Gamma)$ indicates that the realized gates inside selected AGNN layers are either kept unchanged or replaced by a perturbed gate map during the forward pass.
Let $\mathcal{P}_{\mathcal{S}_{\mathrm{pisg}}}$ denote a perturbation operator that weakens gates in the layer set $\mathcal{S}_{\mathrm{pisg}}$ while leaving other layers unchanged.
Here $\mathbf{p}$ denotes pre-sigmoid/pre-softmax logits (not probabilities).
All PISG operations are performed in logit space.
At each denoising step, PISG evaluates two parameter-sharing branches:
\[
\mathbf{p}_t^{\text{normal}} = F_\theta(\mathbf{x}_t, t, G;\, \Gamma_t), \qquad
\mathbf{p}_t^{\text{perturbed}} = F_\theta(\mathbf{x}_t, t, G;\, \mathcal{P}_{\mathcal{S}_{\mathrm{pisg}}}(\Gamma_t)),
\]
where $\Gamma_t$ denotes the gate map realized at step $t$.
The perturbed branch degrades only the gate-mediated structural selection inside the denoiser; it does not alter the instance graph, the noisy state, the task objective, or the constrained decoder.
The guided prediction extrapolates along the direction from perturbed to normal:
\begin{equation}
\label{eq:pisg}
\mathbf{p}_t^{\text{guided}}
= \mathbf{p}_t^{\text{normal}} + s_t \, \boldsymbol{\delta}_t
= \mathbf{p}_t^{\text{normal}} + s_t \bigl(\mathbf{p}_t^{\text{normal}} - \mathbf{p}_t^{\text{perturbed}}\bigr),
\end{equation}
where $s_t$ follows a guidance schedule (Section~\ref{sec:method:implementation}).
We call $\boldsymbol{\delta}_t = \mathbf{p}_t^{\text{normal}} - \mathbf{p}_t^{\text{perturbed}}$ the \emph{PISG guidance signal}.
It measures how the current prediction changes when structure-sensitive gate selectivity is weakened.
The guided logits $\mathbf{p}_t^{\text{guided}}$ replace the normal logits in reverse sampling and in the final candidate ranking before constrained decoding; the decoder's feasibility checks themselves are unchanged.

The guidance signal $\boldsymbol{\delta}_t$ is a finite intervention on gate selectivity, not an injection of external constraint knowledge.
Although the perturbation operator itself is task-agnostic, any task-specific pattern in $\boldsymbol{\delta}_t$ must arise from the trained denoiser's gate pathway rather than from externally injected information.
This motivates a testable prediction: if the denoiser's gate sensitivity is constraint-aligned, $\boldsymbol{\delta}_t$ should concentrate at task-specific constraint boundaries rather than uniformly rescaling logits (a local-linearization derivation supporting this view appears in Appendix~\ref{app:diagnostic-derivation}).

\subsection{Perturbation Operators and Inference}
\label{sec:method:implementation}

Equation~\eqref{eq:pisg} leaves three design choices open: how to construct the perturbed gate map $\mathcal{P}_{\mathcal{S}_{\mathrm{pisg}}}(\Gamma_t)$, how to schedule the guidance strength $s_t$, and which layers to perturb through $\mathcal{S}_{\mathrm{pisg}}$.
The perturbed branch should be degraded specifically along the instance-structure pathway, preserving all other pipeline components.
This keeps the normal-minus-perturbed direction interpretable as a structure-sensitive signal before decoding.

\paragraph{Perturbation operators.}
For layers $\ell\in\mathcal{S}_{\mathrm{pisg}}$, the perturbation operator replaces each gate vector $\mathbf{g}_{ij}^{(\ell)}$ by a spatially non-selective surrogate:
\begin{equation}
\label{eq:perturbation}
\mathcal{P}_{\mathcal{S}_{\mathrm{pisg}}}(\Gamma_t):
\quad
\mathbf{g}_{ij}^{(\ell)}
\mapsto
\begin{cases}
c\cdot \mathbf{1} & \text{(\texttt{uniform})}, \\[3pt]
\mu_t^{(\ell)}\cdot \mathbf{1} & \text{(\texttt{adaptive})},
\end{cases}
\qquad
\ell\in\mathcal{S}_{\mathrm{pisg}},
\end{equation}
where \(c=\tfrac{1}{2}\) by default and
\(\mu_t^{(\ell)} = \frac{1}{|E|}\sum_{(i',j')\in E} \mathrm{mean}(\mathbf{g}_{i'j'}^{(\ell)})\)
averages over all edges and hidden channels for the current instance.
Both operators remove edge-wise spatial selectivity, i.e., which pairwise relations receive stronger message passing.
They differ only in whether the replacement magnitude is fixed globally (\texttt{uniform}) or adapted to the current instance and layer (\texttt{adaptive}).
An additional strategy (random-gate) is compared in Appendix~\ref{app:method-details}.

\paragraph{Guidance schedule.}
The guidance schedule decomposes the per-step strength into a global scale $s$ and a temporal weight $w(t)$:
\begin{equation}
\label{eq:schedule}
s_t = s \cdot w(t), \qquad
w(t) =
\begin{cases}
1 & \text{(\texttt{const})}, \\[3pt]
\dfrac{T_s - t}{T_s - 1} & \text{(\texttt{linear},\; $T_s > 1$)}.
\end{cases}
\end{equation}
Here $t$ indexes the denoising step ($t = T_s, \ldots, 1$), distinct from the underlying diffusion timestep.
Empirically, constant scheduling works best for MIS and linear for TSP (Section~\ref{sec:exp:sensitivity}).

\paragraph{Layer selection.}
The perturbation need not span all GNN layers.
A layer set $\mathcal{S}_{\text{pisg}} \subseteq \{1,\dots,L\}$ restricts which layers have their gating replaced.
In our 12-layer backbones, middle-to-late layers (roughly layers 5--9) form a broad safe zone where most configurations improve over the baseline.
The sensitivity analysis is summarized in Section~\ref{sec:exp:sensitivity} and detailed in Appendix~\ref{app:ablation}.

These choices control where and how strongly the same structural discrepancy is amplified.
They do not introduce objective information or modify feasibility checks; their role is to expose prediction--constraint alignment already present in the trained denoiser.
The two branches share all parameters and can be evaluated by batch doubling, with pseudocode in Algorithm~\ref{alg:pisg} (Appendix~\ref{app:method-details}).
Section~\ref{sec:signal-characterization} empirically tests the resulting directional structure, and Section~\ref{sec:experiments} evaluates PISG across all four solvers, with overhead measurements in Section~\ref{sec:exp:efficiency}.
Edge-gating perturbation is one instance of a procedure: identify the structure pathway through which instance structure enters the prediction, degrade it, and extrapolate; Appendix~\ref{app:gensco} gives a second instance, attention-bias perturbation in a Transformer solver.

%% file: sec/signal_characterization.tex
\section{Signal Analysis: Constraint-Boundary Discriminability}
\label{sec:signal-characterization}

Section~\ref{sec:method} motivates a testable prediction: $\boldsymbol{\delta}_t$ should concentrate at task-specific constraint boundaries rather than rescaling predictions uniformly.
We test this through three diagnostics: local degree-boundary alignment in TSP, conflict-boundary separation in MIS, and a temperature-scaling negative control (reported in Appendix~\ref{app:directional-vs-temperature}).
All results aggregate over 5 seeds (41--45, matching the main tables). PISG extrapolates along the logit-space $\boldsymbol{\delta}_t$ (Equation~\eqref{eq:pisg}); because decoding acts on probabilities, the diagnostics measure its probability-space counterpart $\tilde{\boldsymbol{\delta}}_t = \psi(\mathbf{p}_t^{\text{normal}}) - \psi(\mathbf{p}_t^{\text{perturbed}})$, where $\psi$ is the solver's standard logit-to-probability transform, and rank candidates by normal-branch probability.

\subsection{TSP: Degree-2 Split}
\label{sec:signal:tsp}

A valid TSP tour requires every node to connect to exactly two selected edges:
\begin{equation}
\label{eq:tsp}
\textstyle\sum_{j:(i,j)\in E} x_{ij} = 2 \quad \forall\, i \in V.
\end{equation}
We focus on the degree-2 boundary as the local constraint boundary most directly exposed in edge rankings; global subtour elimination is not directly measured by this diagnostic.
Under the denoiser's current ranking, the top-two incident edges form the local degree-compatible candidate set.
If $\tilde{\boldsymbol{\delta}}_t$ reflects improved degree-boundary discriminability, it should reinforce these two candidates and suppress lower-ranked incident edges, since accepting additional incident edges would breach the degree-2 limit.

\begin{wrapfigure}{r}{0.35\linewidth}
\centering
\includegraphics[width=\linewidth]{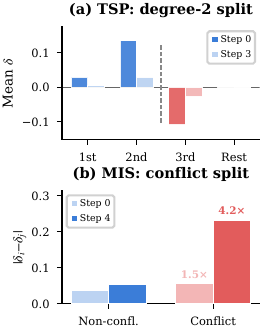}
\caption{Probability-space signal analysis ($\tilde{\boldsymbol{\delta}}_t$).
(a)~TSP: positive for top-2 edges, negative beyond (degree-2 split).
(b)~MIS: conflict-edge endpoint separation is $4.2\times$ non-conflict at step~4.}
\label{fig:signal-characterization}
\end{wrapfigure}

We rank each node's incident edges by normal-branch probability and compute mean $\tilde{\boldsymbol{\delta}}_t$ per rank position across all nodes.
Figure~\ref{fig:signal-characterization}(a) confirms the predicted pattern: at step~3, mean $\tilde{\boldsymbol{\delta}}_t$ is positive for the top-two ranked edges and negative from the third-ranked edge onward; the sign reversal falls precisely at the degree-2 boundary.
This indicates that PISG sharpens competition at the constraint boundary rather than uniformly amplifying the current leader.
The same directional pattern persists across denoising steps with varying magnitude (Appendix~\ref{app:tsp-signal-detail}).

\subsection{MIS: Conflict-Edge Split}
\label{sec:signal:mis}

MIS imposes a structurally different constraint.
Two adjacent nodes cannot both belong to an independent set.
\begin{equation}
\label{eq:mis}
x_i + x_j \le 1 \quad \forall\, (i,j) \in E.
\end{equation}
We call an edge a \emph{predicted conflict edge} when both endpoints have transformed normal-branch scores above $\vartheta{=}0.3$; these are the decisions where the decoder must choose one endpoint over the other.
If $\tilde{\boldsymbol{\delta}}_t$ reflects improved conflict-boundary discriminability, it should pull apart the scores of conflicting endpoints.

Unlike TSP, where the signal manifests as a sign reversal, MIS $\tilde{\boldsymbol{\delta}}_t$ is predominantly positive for high-score nodes.
The constraint signal instead appears as endpoint-separation asymmetry between predicted conflict and non-conflict edges.
Figure~\ref{fig:signal-characterization}(b) shows this clearly.
The endpoint separation $|\tilde{\delta}_i - \tilde{\delta}_j|$ on conflict edges is $1.5\times$ that of non-conflict edges at step~0 and grows to $4.2\times$ at step~4.
This growing asymmetry shows that the guidance signal concentrates on the most contested constraint decisions as predictions sharpen, giving the greedy decoder a stronger cue for which node to retain (threshold robustness in Appendix~\ref{app:mis-signal-detail}).

Together, the degree-2 sign reversal in TSP and the conflict-edge asymmetry in MIS show that the guidance direction improves constraint-boundary discriminability in task-specific ways, supporting the claim that PISG performs directional, constraint-aligned signal amplification rather than uniform score sharpening.
A temperature-scaling negative control confirms that scalar rescaling cannot produce these patterns (Appendix~\ref{app:directional-vs-temperature}).
Cross-solver replication and trajectory-level probes appear in Appendices~\ref{app:mechanism-cross-method}--\ref{app:stepwise-decode}.
These diagnostics characterize the geometry of the guidance direction rather than per-instance benefit.
After controlling for baseline quality, the degree-2 sign-compliance rate (sc@2; Appendix~\ref{app:tsp-signal-detail}) no longer tracks final TSP quality, whereas the conflict-edge magnitude asymmetry (CMA; Appendix~\ref{app:mis-signal-detail}) retains an association with final MIS quality; neither association is causal (Appendix~\ref{app:diagnostics-quality}).
The TSP evidence concerns the local degree-2 boundary: a fragment-closing probe finds no subtour-level signal (Appendix~\ref{app:subtour}).

%% file: sec/experiments.tex
\section{Experiments}
\label{sec:experiments}

We next ask whether the constraint-boundary discriminability observed at the signal level translates into better feasible solutions after constrained decoding.
The primary evaluation therefore disables downstream objective-guided refinement (\(T_g{=}0\)) and local search, so that the comparison focuses on the predictions passed to constrained decoding.

We evaluate PISG on all four solvers of the edge-gated AGNN family (Section~\ref{sec:related}): DIFUSCO~\citep{sun2023difusco} (\(T_s{=}100\)), T2T~\citep{li2023t2t} (\(T_s{=}50\)), FastT2T~\citep{li2024fastt2t} (\(T_s{=}5\)), and StruDiCO~\citep{wang2025strudico} (\(T_s{=}5\)).
Benchmarks cover TSP at four scales (50--1{,}000 nodes; gap\% to Concorde~\citep{applegate2006concorde}, lower is better) and MIS on two synthetic distributions: sparse Erd\H{o}s--R\'enyi (ER) graphs and denser RB-model graphs (solved cost, higher is better).
The controlled-isolation results report 5-seed means under \(T_g{=}0\), greedy decoding (StruDiCO retains its native decoders), and no local search; Tables~\ref{tab:tsp}--\ref{tab:mis_main} additionally include \(T_g{>}0\) rows to evaluate interaction with objective-side refinement.
The \(T_g{=}0\) setting is an identification design, not a deployment recommendation.
Section~\ref{sec:exp:deployment} then evaluates the deployment pipeline with objective rewrite and 2-opt, and a preliminary transfer to a Transformer solver.
Full details appear in Appendix~\ref{app:full-results}.

\begin{table}[t]
\caption{TSP results across four solvers and four scales.
Gap\% = (cost $-$ Concorde) / Concorde $\times$ 100.
\textbf{Bold}: best gap within each solver block and scale.}
\label{tab:tsp}
\centering\scriptsize
\setlength{\tabcolsep}{2.0pt}
\resizebox{\linewidth}{!}{%
\begin{tabular}{ll ccc ccc ccc ccc}
\toprule
 & & \multicolumn{3}{c}{TSP-50} & \multicolumn{3}{c}{TSP-100} & \multicolumn{3}{c}{TSP-500} & \multicolumn{3}{c}{TSP-1000} \\
\cmidrule(lr){3-5}\cmidrule(lr){6-8}\cmidrule(lr){9-11}\cmidrule(l){12-14}
Method & \(T_s/T_g\) & Cost & Gap\% & Time & Cost & Gap\% & Time & Cost & Gap\% & Time & Cost & Gap\% & Time \\
\midrule
Concorde & -- & 5.688 & 0.000 & -- & 7.756 & 0.000 & -- & 16.55 & 0.00 & -- & 23.12 & 0.00 & -- \\
\midrule
DIFUSCO & 100/0 & 5.708 & 0.367 & 6.2m & 7.864 & 1.388 & 6.7m & 18.52 & 11.91 & 3.8m & 26.19 & 13.31 & 9.6m \\
\;+PISG & 100/0 & 5.699 & \textbf{0.203} & 6.2m & 7.800 & \textbf{0.567} & 10.2m & 18.11 & \textbf{9.44} & 5.8m & 25.65 & \textbf{10.95} & 16.7m \\
\midrule
T2T & 50/0 & 5.727 & 0.683 & 3.5m & 7.882 & 1.624 & 3.6m & 18.42 & 11.33 & 76s & 26.13 & 13.05 & 2.4m \\
\;+PISG & 50/0 & 5.719 & \textbf{0.543} & 3.5m & 7.815 & \textbf{0.758} & 5.0m & 18.18 & \textbf{9.88} & 96s & 26.09 & \textbf{12.87} & 3.6m \\
\midrule
\multirow{2}{*}{FastT2T} & 5/0 & 5.689 & 0.023 & 45s & 7.762 & 0.079 & 52s & 17.51 & 5.84 & 32s & 24.61 & 6.46 & 48s \\
 & 5/5 & 5.688 & \textbf{0.0069} & 135s & 7.758 & 0.025 & 154s & 16.92 & 2.27 & 68s & 23.99 & 3.78 & 177s \\
\multirow{2}{*}{\;+PISG} & 5/0 & 5.689 & 0.017 & 48s & 7.760 & 0.051 & 55s & 17.40 & 5.15 & 33s & 24.49 & 5.93 & 56s \\
 & 5/5 & 5.688 & 0.0071 & 141s & 7.758 & \textbf{0.023} & 154s & 16.87 & \textbf{1.94} & 69s & 23.86 & \textbf{3.22} & 182s \\
\midrule
\multirow{2}{*}{StruDiCO} & 5/0 & 5.689 & 0.027 & 45s & 7.770 & 0.187 & 46s & 17.34 & 4.81 & 33s & 25.83 & 11.75 & 46s \\
 & 5/5 & 5.688 & 0.0125 & 71s & 7.758 & \textbf{0.031} & 77s & 17.00 & 2.71 & 42s & 25.11 & 8.60 & 74s \\
\multirow{2}{*}{\;+PISG} & 5/0 & 5.689 & 0.025 & 48s & 7.770 & 0.179 & 53s & 17.33 & 4.74 & 37s & 25.43 & 9.99 & 58s \\
 & 5/5 & 5.688 & \textbf{0.0123} & 70s & 7.758 & 0.031 & 82s & 16.95 & \textbf{2.45} & 46s & 24.67 & \textbf{6.72} & 86s \\
\bottomrule
\end{tabular}
}
\end{table}

\subsection{Main Results under Controlled Isolation}
\label{sec:exp:main}

\begin{wraptable}{r}{0.46\linewidth}
\caption{MIS results on ER and RB.
Cost is higher-better; gap\% is lower-better and negative means exceeding the reference.
\textbf{Bold}: best within each solver block.}
\label{tab:mis_main}
\centering\scriptsize
\setlength{\tabcolsep}{1.2pt}
\begin{tabular}{ll ccc ccc}
\toprule
    & & \multicolumn{3}{c}{ER} & \multicolumn{3}{c}{RB} \\
\cmidrule(lr){3-5}\cmidrule(l){6-8}
Method & \(T_s/T_g\) & Cost$\uparrow$ & Gap\%$\downarrow$ & Time & Cost$\uparrow$ & Gap\%$\downarrow$ & Time \\
\midrule
Reference & -- & 41.35 & 0.00 & --- & 20.09 & 0.00 & --- \\
\midrule
DIFUSCO       & 100/0 & 40.63 &  1.73 & 4.0m & ---   & ---  & --- \\
\;+PISG       & 100/0 & 42.55 & $-$\textbf{2.91} & 6.5m & ---   & ---  & --- \\
\midrule
T2T & 50/0  & 36.50 & 11.74 & 1.5m & ---   & ---  & --- \\
\;+PISG & 50/0  & 38.25 &  \textbf{7.50} & 2.3m & ---   & ---  & --- \\
\midrule
\multirow{2}{*}{FastT2T} & 5/0 & 37.65 &  8.92 & 33s  & 18.81 & 6.38 & 38s \\
                            & 5/5 & 40.98 &  0.87 & 62s & 19.47 & 3.07 & 93s \\
\multirow{2}{*}{\;+PISG} & 5/0 & 38.89 &  5.92 & 38s  & 18.89 & 5.98 & 41s \\
                            & 5/5 & 41.86 & $-$\textbf{1.25} & 67s & \textbf{19.49} & \textbf{2.98} & 96s \\
\midrule
\multirow{2}{*}{StruDiCO} & 5/0 & 39.89 &  3.51 & 35s  & 19.30 & 3.93 & 48s \\
                            & 5/5 & 42.46 & $-$2.72 & 49s  & 19.74 & 1.77 & 73s \\
\multirow{2}{*}{\;+PISG}  & 5/0 & 40.05 &  3.11 & 41s  & 19.31 & 3.90 & 50s \\
                            & 5/5 & 42.55 & $-$\textbf{2.93} & 58s  & 19.75 & \textbf{1.72} & 79s \\
\bottomrule
\end{tabular}
\end{wraptable}

Table~\ref{tab:tsp}\label{sec:exp:tsp} reports TSP results.
In the controlled \(T_g{=}0\) rows, PISG improves all 16 solver--scale combinations.
The largest absolute reductions occur in solvers with the largest residual headroom, notably DIFUSCO and T2T: on TSP-100, DIFUSCO gap drops from \(1.39\%\) to \(0.57\%\), and T2T drops from \(1.62\%\) to \(0.76\%\).
Even near-optimal short-trajectory baselines improve: FastT2T gap on TSP-100 falls from \(0.079\%\) to \(0.051\%\).

Table~\ref{tab:mis_main}\label{sec:exp:mis} reports MIS results.
On ER, all four solvers improve: DIFUSCO gains \(+4.7\%\) in solved cost and obtains a negative gap relative to the reported reference; T2T and FastT2T reduce their gaps by \(4.2\) and \(3.0\) percentage points, respectively.
On RB, where only FastT2T and StruDiCO have pretrained checkpoints, FastT2T improves while StruDiCO is effectively neutral.
Per-seed standard deviations and statistical significance tests are reported in Appendix~\ref{app:stat}.
With five seeds, 16 of the 22 comparisons reach \(p{<}0.05\); in a 20-seed follow-up of the remaining six, four reach \(p{<}0.05\) (T2T TSP-1000 and StruDiCO TSP-100/500/MIS-ER), StruDiCO TSP-50 remains inconclusive, and StruDiCO MIS-RB is effectively null.

On TSP, absolute gains scale with residual headroom: solvers with larger baseline gaps (DIFUSCO, T2T) benefit most. On both tasks, StruDiCO---which already exploits structure aggressively---shows smaller improvements that require lower guidance strength (Section~\ref{sec:exp:sensitivity}).

\paragraph{Hyperparameter sensitivity.}
\label{sec:exp:sensitivity}
PISG is training-free but not tuning-free.
Family defaults are identified from a full sensitivity analysis on FastT2T (Appendix~\ref{app:val-protocol}) and checked on held-out validation splits; they transfer to DIFUSCO and T2T without re-tuning.
The selected defaults are: \texttt{adaptive}/\(s{=}2.0\)/L7/\texttt{linear} for TSP, \texttt{uniform}/\(s{=}2.0\)/L7/\texttt{constant} for MIS-ER, and \texttt{adaptive}/\(s{=}2.0\)/L7/\texttt{constant} for MIS-RB. On the validation splits, the MIS-ER default is also the best configuration, while the TSP-100 default improves over the baseline; the TSP default also transfers across scales.
Full sweeps in Appendix~\ref{app:ablation} show a broad safe region: on TSP-100, \(82\%\) of 180 gate-level configurations improve over the baseline and all configurations in the central region (scale~$\in[1,3]$, layers~$\in[5,9]$) improve.
StruDiCO is a systematic exception, requiring guidance strength an order of magnitude lower (\(s{=}0.1\)--\(0.2\)), consistent with its stronger built-in structure utilization (Appendix~\ref{app:config-selection}).
Frozen configurations transfer within the coverage of the training distribution but fail at a high-density MIS endpoint, in most extrapolations to larger TSP scales, and on large TSPLIB instances (Appendix~\ref{app:shift-transfer}).

\subsection{Deployment Pipeline and Transfer}
\label{sec:exp:deployment}

\paragraph{Objective-side refinement.}
\label{sec:exp:practical}%
The \(T_g{>}0\) rows in Tables~\ref{tab:tsp}--\ref{tab:mis_main} combine PISG with objective-side refinement for FastT2T and StruDiCO.
The interaction is headroom-dependent: at larger scales (TSP-500/1000), PISG+\(T_g\) consistently improves over baseline+\(T_g\); at smaller scales where the baseline already approaches optimality, the combination is near-neutral.
On MIS-ER, PISG+\(T_g\) improves throughout: FastT2T at \(T_g{=}5\) obtains a negative gap relative to the reported reference (\({-}1.25\%\)).
This supports the pipeline separation introduced in the Introduction: PISG acts before constrained decoding by improving prediction--constraint alignment, whereas \(T_g\) acts through objective-side refinement.
Their complementarity therefore depends on residual headroom rather than on redundant score sharpening.

\paragraph{Full pipeline with 2-opt.}
The settings above omit local search.
In the full pipeline with \(T_g{=}5\) and 2-opt run to convergence, the frozen configurations retain a positive mean effect in three of four large-scale cells (Appendix~\ref{app:deployment}).
On FastT2T, PISG removes \(6.1\%\) and \(4.5\%\) of the remaining gap on TSP-500 and TSP-1000; StruDiCO TSP-500 retains a small gain, whereas StruDiCO TSP-1000 reverses at the converged 2-opt endpoint.
These gains are much smaller than under controlled isolation, consistent with downstream refinement absorbing part of the residual headroom.

\paragraph{Preliminary transfer to a Transformer solver.}
We also apply PISG to GenSCO~\citep{li2025gensco}, a Transformer solver in which instance structure enters through an adjacency bias on attention logits rather than through edge gating.
The structure-degraded branch weakens this attention-bias pathway at selected layers, and configurations are frozen across search budgets (Appendix~\ref{app:gensco}).
On TSP, PISG yields positive mean effects in 10 of 12 cells, including all six TSP-1000 cells, while TSP-500 with in-loop 2-opt reverses at two of three budgets. On MIS, gains appear only on ER at the two smaller budgets and only together with a higher search flip rate; RB shows no measurable effect.
These results are preliminary evidence from one Transformer solver with five seeds.

\subsection{Compute--Quality Trade-off}
\label{sec:exp:efficiency}

\begin{figure}[t]
\centering
\includegraphics[width=0.95\linewidth]{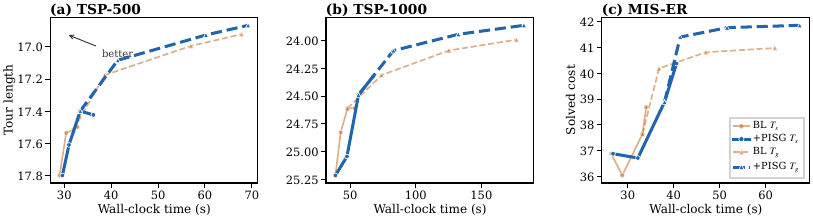}
\caption{Compute--quality trade-off on FastT2T.
Solid: $T_s$ scaling ($T_g{=}0$); dashed: $T_g$ scaling ($T_s{=}5$).
PISG shifts the frontier in later-budget regimes and adds 3--17\% wall-clock overhead (Appendix~\ref{app:efficiency}).}
\label{fig:pareto}
\end{figure}

PISG adds one perturbed forward pass per denoising step; the normal and perturbed branches are batched.
At \(T_g{=}0\), this adds 3--17\% wall-clock overhead on FastT2T and 4--26\% on StruDiCO (5-seed mean across all tasks; Appendix~\ref{app:efficiency}).
Thus PISG is a targeted pre-decoding alignment step with moderate but nonzero inference cost.

Figure~\ref{fig:pareto} uses FastT2T as a representative short-trajectory testbed that enables dense \(T_s\) and \(T_g\) sweeps.
Along the \(T_s\) axis (solid lines), PISG improves over the baseline from \(T_s{=}5\) onward (from \(T_s{=}3\) on MIS-ER).
On large-scale TSP (500/1000), PISG at \(T_s{=}5\) achieves lower gap than the baseline at \(T_s{=}10\) at comparable or lower wall-clock cost (Appendix~\ref{app:compute-equiv}).
On TSP-50/100 it beats only the closest-time baseline, and on both MIS distributions adding denoising steps is more effective than PISG.
Along the \(T_g\) axis (dashed lines), PISG+\(T_g\) reaches or exceeds baseline+\(T_g\) at higher rewrite budgets on TSP-500 and TSP-1000.
On MIS-ER, PISG+\(T_g\) improves throughout the tested range, though early budget points overlap.

%% file: sec/conclusion.tex
\section{Conclusion and Limitations}
\label{sec:conclusion}

We introduced PISG, a training-free denoiser-internal guidance mechanism that reveals and amplifies constraint-aligned sensitivity in edge-gated AGNN denoisers.
Signal analysis shows that task-agnostic edge-gating perturbations expose task-specific constraint-boundary discriminability---a degree-2 boundary split in TSP and conflict-edge endpoint separation in MIS---rather than uniform score rescaling.
Under controlled isolation, PISG improves every tested solver--scale combination on TSP and every solver on MIS-ER (solved cost up to $+4.8\%$).
Constraint-aligned signal amplification and objective-side refinement ($T_g$) are mechanistically distinct and non-redundant; where residual headroom remains, their combination yields further improvement.
After 2-opt, the benefit persists in three of four large-scale cells, and attention-bias perturbation gives a preliminary transfer to the Transformer-based GenSCO.

Several limitations apply.
The extra forward pass at each denoising step raises peak GPU memory, and the method offers limited value when $T_s$ is small---on MIS, allocating the same compute to additional denoising steps is more effective.
PISG is training-free but not tuning-free: a validation sweep (${\sim}$50 configurations) is needed for each task--distribution pair, and frozen configurations can fail under larger density or scale shifts.
The constraint-aligned interpretation rests on empirical, partly correlational evidence; the TSP diagnostics cover only the local degree-2 boundary and lack subtour-level signal, while per-instance predictive power of the proposed metrics remains limited (Appendices~\ref{app:diagnostics-quality}--\ref{app:subtour}).
Finally, edge-gating and attention-bias perturbation are two instances of degrading a structure pathway; extending to other backbones, constraint systems, or objectives the model has not been trained on would require further investigation.

%% file: appendix/appendix.tex
\noindent\textbf{Seed convention.} Unless otherwise noted, appendix results report 5-seed means (seeds 41--45). Signal characterization in the main text (Section~\ref{sec:signal-characterization}) also uses 5-seed aggregates. Single-seed analyses in ablation, trajectory probes, and cross-solver replication experiments are marked at first occurrence.

\section{Method Details and Transfer to a Transformer Solver}
\label{app:method-details}

This appendix provides implementation details for reproducibility, supporting Section~\ref{sec:method}. It includes the PISG inference pseudocode, the full set of perturbation operator definitions, and an empirical comparison of perturbation strategies that validates the design rationale. Appendix~\ref{app:gensco} applies the same procedure to GenSCO, a Transformer solver without edge gating, and reports its configuration, results, and signal diagnostics.

\subsection{PISG Inference Pseudocode}

Algorithm~\ref{alg:pisg} summarizes the complete PISG inference procedure described in Section~\ref{sec:method:implementation}.

\begin{algorithm}[H]
\small
\caption{PISG Inference}\label{alg:pisg}
\begin{algorithmic}[1]
\REQUIRE \parbox[t]{0.85\linewidth}{trained denoiser $F_\theta$; instance graph $G$; perturbation operator $\mathcal{P}$; guidance strength $s$; step schedule $w(\cdot)$; layer set $\mathcal{S}_{\text{pisg}}$; sampling steps $T_s$}
\ENSURE final guided logits $\mathbf{p}_1^{\text{guided}}$, decoded to feasible $\hat{\mathbf{x}}$ by greedy/neighbor-aware decoder
\STATE $\mathbf{x}_{T_s} \sim$ initial noise distribution
\FOR{$t = T_s, T_s{-}1, \ldots, 1$}
    \STATE $\mathbf{p}_t^{\text{normal}} \leftarrow F_\theta(\mathbf{x}_t, t, G;\, \Gamma_t)$
    \STATE $\mathbf{p}_t^{\text{perturbed}} \leftarrow F_\theta(\mathbf{x}_t, t, G;\, \mathcal{P}_{\mathcal{S}_{\text{pisg}}}(\Gamma_t))$
    \STATE $s_t \leftarrow s \cdot w(t)$
    \STATE $\mathbf{p}_t^{\text{guided}} \leftarrow \mathbf{p}_t^{\text{normal}} + s_t(\mathbf{p}_t^{\text{normal}} - \mathbf{p}_t^{\text{perturbed}})$
    \STATE $\mathbf{x}_{t-1} \leftarrow \mathrm{Sample}(\mathbf{p}_t^{\text{guided}}, t)$
\ENDFOR
\RETURN $\hat{\mathbf{x}} = \mathrm{Decoder}(\mathbf{p}_1^{\text{guided}})$
\end{algorithmic}
\end{algorithm}

\subsection{Perturbation Operator Definitions}
\label{app:perturbation-ops}

Section~\ref{sec:method:implementation} introduces two representative operators (\texttt{uniform} and \texttt{adaptive}). Here we define the complete set, including one additional strategy used in the ablation study (Table~\ref{tab:perturbation}).

All three gate-targeting operators replace the \(d_h\)-dimensional gate vector \(\mathbf{g}_{ij}^{(\ell)}\) at each target layer \(\ell \in \mathcal{S}_{\text{pisg}}\):
\begin{equation}
\label{eq:perturbation-full}
\mathcal{P}_{\mathcal{S}_{\mathrm{pisg}}}(\Gamma): \quad \mathbf{g}_{ij}^{(\ell)} \mapsto
\begin{cases}
c \cdot \mathbf{1} & \text{(\texttt{uniform})}, \\[3pt]
\mu_t^{(\ell)} \cdot \mathbf{1} & \text{(\texttt{adaptive})}, \\[3pt]
\mathbf{u}_{ij}^{(\ell)}, \ \mathbf{u}_{ij}^{(\ell)} \sim \mathcal{U}(0,1)^{d_h} & \text{(\texttt{random-gate})},
\end{cases}
\end{equation}
where \(c = \tfrac{1}{2}\) by default and \(\mu_t^{(\ell)} = \frac{1}{|\mathcal{E}|}\sum_{(i',j') \in \mathcal{E}} \mathrm{mean}(\mathbf{g}_{i'j'}^{(\ell)})\) averages over all edges and hidden channels for the current instance. \texttt{Uniform} is an instance-agnostic constant that completely severs the link between instance structure and gating weights. \texttt{Adaptive} preserves instance-level activation magnitude but removes spatial selectivity. \texttt{Random-gate} replaces each gate vector with i.i.d.\ uniform noise in the gate range \((0,1)\); it destroys the edge--gate correspondence but does not preserve the learned gate values.

\paragraph{Non-gate perturbation.}
A fourth baseline, \texttt{input-noise}, does not target the gating mechanism. Instead, it adds Gaussian noise to the (scaled) noisy solution state \(\mathbf{x}_t\) fed to the perturbed branch, while the gates are computed as usual without replacement:
\begin{equation}
\label{eq:perturbation-inputnoise}
\mathcal{P}_{\texttt{inp}}:\quad \mathbf{x}_t \;\mapsto\; \mathbf{x}_t + \boldsymbol{\epsilon}, \qquad \boldsymbol{\epsilon} \sim \mathcal{N}(0,\, \sigma^2 \mathbf{I}),
\end{equation}
where \(\sigma = 0.5\). Unlike the three gate-targeting operators in Eq.~\eqref{eq:perturbation-full}, \(\mathcal{P}_{\texttt{inp}}\) degrades the input to the denoiser without replacing the structure-sensitive gates. It is included in Table~\ref{tab:perturbation} as a control that tests whether effective guidance requires targeting the gate mechanism rather than degrading information generically.

\subsection{Diagnostic Interpretation of the Guidance Signal}
\label{app:diagnostic-derivation}

View the selected gate vectors as a concatenated continuous variable and define $\Delta\Gamma_t = \Gamma_t - \mathcal{P}_{\mathcal{S}_{\mathrm{pisg}}}(\Gamma_t)$.
A local linearization of $F_\theta$ with respect to this gate variable gives the diagnostic approximation
\[
\boldsymbol{\delta}_t
= F_\theta(\mathbf{x}_t,t,G;\Gamma_t) - F_\theta(\mathbf{x}_t,t,G;\mathcal{P}_{\mathcal{S}_{\mathrm{pisg}}}(\Gamma_t))
\approx J_{\theta,t}^{\Gamma}\,\Delta\Gamma_t,
\]
where $J_{\theta,t}^{\Gamma}$ is the denoiser's Jacobian with respect to the selected gate map.
This approximation is used only as a diagnostic lens, since $\mathcal{P}_{\mathcal{S}_{\mathrm{pisg}}}$ is a finite intervention rather than an infinitesimal perturbation.
Nevertheless, it clarifies that $\boldsymbol{\delta}_t$ probes the denoiser's local sensitivity to gate-mediated structural selection ($J_{\theta,t}^{\Gamma}$), projected along the direction removed by the perturbation ($\Delta\Gamma_t$).
If this sensitivity is constraint-aligned---i.e., if the denoiser has learned to route messages preferentially along constraint-relevant edges---then $\boldsymbol{\delta}_t$ should concentrate at task-specific constraint boundaries rather than uniformly rescaling logits.
Section~\ref{sec:signal-characterization} tests this prediction empirically.

\subsection{Why Effective Degradation Must Sever Structure-Sensitive Gating}
\label{app:degradation}

\begin{table*}[t]
\caption{Perturbation strategy comparison on TSP-100 and MIS-ER (FastT2T, \(T_s{=}5\), \(T_g{=}0\), greedy, seed~42). Strategies are grouped into gate-targeting (operating on the gate vector \(\mathbf{g}_{ij}^{(\ell)}\), Eq.~\eqref{eq:perturbation-full}) and non-gate (operating on the denoiser input, Eq.~\eqref{eq:perturbation-inputnoise}).}
\label{tab:perturbation}
\centering
\scriptsize
\begin{tabular*}{\textwidth}{@{\extracolsep{\fill}}ll c ccc c@{}}
\toprule
& & & \multicolumn{3}{c}{Gate-targeting} & Non-gate \\
\cmidrule(lr){4-6}\cmidrule(l){7-7}
Task & Metric & Base & Unif. & Adpt. & Rand-g. & Inp-noise \\
\midrule
TSP & gap\(\downarrow\) & 0.086 & 0.057 & 0.056 & \textbf{0.046} & 0.077 \\
MIS & cost\(\uparrow\) & 37.81 & \textbf{39.30} & 33.48 & \underline{38.58} & 37.81 \\
\bottomrule
\end{tabular*}
\end{table*}

Table~\ref{tab:perturbation} groups the four strategies into \emph{gate-targeting} (operating directly on the gate vector) and \emph{non-gate} (operating on the denoiser input). In this two-task ablation, gate-targeting emerges as the relevant search space for effective perturbation: the two best cross-task strategies (\texttt{uniform}, \texttt{random-gate}) both belong to this family. However, not all gate-targeting strategies are reliable: \texttt{adaptive} severely degrades MIS-ER (\(33.48\) vs.\ baseline \(37.81\)), likely because preserving instance-level activation magnitude produces an impure contrast that amplifies harmful directions. The non-gate baseline (\texttt{input-noise}) yields no improvement on MIS-ER. These results support the view that effective degradation in this setting requires both targeting the gating mechanism \emph{and} breaking edge--instance correspondence. Among the four strategies, \texttt{uniform} and \texttt{random-gate} satisfy both requirements and are the most reliable cross-task choices in this two-task ablation.

\subsection{Adapting PISG to GenSCO}
\label{app:gensco}

This subsection applies PISG to GenSCO~\citep{li2025gensco}, a Transformer-based diffusion solver without edge gating. The procedure is the same as for AGNN denoisers: identify where instance structure enters the prediction, construct a structure-degraded branch there, and extrapolate with Equation~\eqref{eq:pisg}. We report it as preliminary evidence on one Transformer solver with five paired seeds.

\paragraph{Perturbation target.}
GenSCO operates on node embeddings, and instance structure enters through an adjacency bias added to the attention logits. This bias is therefore the perturbation target. For layers \(\ell\in\mathcal{S}_{\text{pisg}}\), the perturbation operator scales down the adjacency bias and, for TSP, also replaces each node's attention weights with a less node-specific average. Like the gate surrogates in Equation~\eqref{eq:perturbation}, these operations remove part of the edge-wise structural selectivity, while the structure-degraded branch shares all inputs and parameters with the normal branch.

\paragraph{Configuration.}
For TSP, one configuration (bias and attention perturbation, \(s{=}0.5\), \(\mathcal{S}_{\text{pisg}}{=}\{3\}\), \texttt{linear} schedule) is fixed for all twelve cells: two scales, three search budgets \(C\in\{10,20,40\}\) cycles, and two settings of GenSCO's in-loop 2-opt (\(0\) or \(20\) steps). For MIS, one bias-only configuration per distribution is selected at the reference budget \(C{=}500\) on a subset of the evaluation graphs and frozen for \(C\in\{1000,2000\}\): ER uses \(s{=}0.93\) with layers~3--6, and RB uses \(s{=}0.25\) with layers~7--12, both with a \texttt{linear} schedule. Both MIS configurations also raise GenSCO's search flip rate from the default \([0.25,0.40]\) to \([0.39,0.54]\) (ER) and \([0.35,0.50]\) (RB). No cell-wise re-tuning is performed.

\begin{table}[t]
\caption{GenSCO TSP results with one frozen configuration (seeds 41--45). Entries are the mean paired gap reduction in percentage points (\(>0\): PISG better) with paired wins out of 5. \(C\): search cycles; 2-opt: GenSCO's in-loop 2-opt steps.}
\label{tab:gensco-tsp}
\centering\scriptsize
\begin{tabular*}{\linewidth}{@{\extracolsep{\fill}}lcccc@{}}
\toprule
Task & 2-opt & \(C{=}10\) & \(C{=}20\) & \(C{=}40\) \\
\midrule
TSP-500 & 0 & \(+\)0.00096 (3/5) & \(+\)0.00106 (5/5) & \(+\)0.00176 (5/5) \\
TSP-500 & 20 & \({-}\)0.00003 (2/5) & \(+\)0.00049 (4/5) & \({-}\)0.00012 (2/5) \\
TSP-1000 & 0 & \(+\)0.04718 (5/5) & \(+\)0.02012 (5/5) & \(+\)0.01427 (5/5) \\
TSP-1000 & 20 & \(+\)0.00399 (5/5) & \(+\)0.00297 (4/5) & \(+\)0.00215 (4/5) \\
\bottomrule
\end{tabular*}
\end{table}

\begin{table}[t]
\caption{GenSCO MIS results with one frozen configuration per distribution (seeds 41--45). Entries are the mean paired change in independent-set size (\(>0\): PISG better) with paired wins out of 5. The baseline is evaluated at both the default and the PISG flip rate, and each row compares against the stronger one (default for ER, PISG flip rate for RB).}
\label{tab:gensco-mis}
\centering\scriptsize
\begin{tabular*}{\linewidth}{@{\extracolsep{\fill}}lccc@{}}
\toprule
Dataset & \(C{=}500\) & \(C{=}1000\) & \(C{=}2000\) \\
\midrule
MIS-ER & \(+\)0.047 (4/5) & \(+\)0.080 (5/5) & \({-}\)0.036 (2/5) \\
MIS-RB & \({-}\)0.003 (2/5) & \(+\)0.004 (4/5) & \({-}\)0.005 (2/5) \\
\bottomrule
\end{tabular*}
\end{table}

\paragraph{Results.}
Table~\ref{tab:gensco-tsp} shows positive mean effects in 10 of the 12 TSP cells. TSP-1000 improves in all six cells with 4--5 paired wins each, including after in-loop 2-opt. TSP-500 improves without 2-opt; with 2-opt the effect depends on the budget and is negative at \(C{=}10\) and \(C{=}40\). On MIS (Table~\ref{tab:gensco-mis}), MIS-ER improves at \(C{=}500\) and \(C{=}1000\) and reverses at \(C{=}2000\). This requires PISG together with the higher flip rate: at the default flip rate, the ER configuration lowers the independent-set size at all three budgets, and the higher flip rate alone also lowers the unguided baseline. MIS-RB shows no measurable effect at either flip rate. With five seeds, the cell-wise results are descriptive; we do not claim per-cell statistical significance.

\paragraph{Signal diagnostics.}
We repeat the diagnostics of Section~\ref{sec:signal-characterization} on GenSCO (seeds 41--45, 128 instances, three search checkpoints). On TSP-100, sc@2 exceeds its within-instance shuffled null at search cycles 0, 4, and 9 by \(0.009\)--\(0.010\) (\(p<10^{-9}\) at each checkpoint). On MIS-ER, the CMA ratio (\(\theta{=}0.3\)) is \(6.5\times\), \(9.5\times\), and \(9.7\times\) at cycles 0, 249, and 499 (\(p<10^{-22}\)). The degree-2 sign pattern and conflict-edge asymmetry therefore also appear in this attention-bias architecture. Their temporal profiles differ from FastT2T, and the diagnostics carry the same boundaries as in Appendices~\ref{app:diagnostics-quality}--\ref{app:subtour}: they characterize the guidance direction rather than predict per-instance benefit.

\paragraph{Scope.}
These results extend the procedure to one solver outside the edge-gated family; broader architectural coverage is left to future work.

\section{Full Experimental Results}
\label{app:full-results}

This appendix provides the complete experimental evidence for Section~\ref{sec:experiments}. It documents the solver implementations, benchmark datasets, evaluation metrics, and seed protocol, and extends the main tables with standard deviations and paired statistical tests (Appendices~\ref{app:full-std}--\ref{app:stat}). These controlled-isolation results constitute the primary empirical evidence for PISG's effectiveness within the edge-gated AGNN family. Appendix~\ref{app:deployment} evaluates PISG in the full deployment pipeline with objective rewrite and 2-opt, and Appendix~\ref{app:efficiency} reports its compute cost and compute-matched comparisons.

\subsection{Experimental Setup}

\paragraph{Solvers.}
We evaluate PISG on four AGNN-based solvers, all sharing the same 12-layer anisotropic GNN backbone with edge gating \citep{bresson2017residual, sun2023difusco} and publicly released checkpoints:
\begin{itemize}
\item \textbf{DIFUSCO} \citep{sun2023difusco}: categorical discrete diffusion, \(T_s{=}100\), greedy decoding. Baseline and PISG both feed the denoiser the training-time state encoding (\(2x{-}1\) with small multiplicative jitter) on TSP and MIS, whereas the official inference code feeds raw \(\{0,1\}\) states; with raw MIS inputs, the default configuration (selected under the training encoding) changes MIS-ER cost by \({-}1.7\%\).
\item \textbf{T2T} \citep{li2023t2t}: discrete diffusion, \(T_s{=}50\), greedy decoding; supports objective-guided refinement (disabled in \(T_g{=}0\) rows).
\item \textbf{FastT2T} \citep{li2024fastt2t}: consistency-based, \(T_s{=}5\), greedy decoding.
\item \textbf{StruDiCO} \citep{wang2025strudico}: structured diffusion with variable-absorption noising, \(T_s{=}5\); its native decoders are retained in all rows including $T_g{=}0$. For MIS, neighbor-aware decoding is a single-pass, deterministic variant that reranks candidates by $p_i / (\epsilon + \sum_{j \in \mathcal{N}(i)} p_j)$, distinct from iterative rewrite. For TSP, a threshold-merge decoder first merges edges whose scores exceed a threshold and then stitches the remaining subpaths in a seeded random order. Threshold \(m_1\) applies to intermediate denoising samples and the initial decoding, and \(m_2\) to rewrite sampling and decoding (active only when \(T_g{>}0\)). Baseline and PISG share \(m_1{=}m_2{=}0.9\) on TSP-50 and \(0.8\) on TSP-100; on TSP-500 and TSP-1000 the baseline uses \(0.7\) and \(0.8\), respectively, and PISG uses \(0.6\) for both, including the deployment pipeline (Appendix~\ref{app:deployment}).
\end{itemize}
All main-table results report mean over 5 independent seeds (41--45). No post-hoc local search (no 2-opt) is applied unless stated (Appendix~\ref{app:deployment}).

\paragraph{Datasets and metrics.}
TSP is evaluated at four scales (50/100/500/1000 nodes; 1{,}280 instances for 50/100, 128 for 500/1000) against Concorde~\citep{applegate2006concorde}, reporting gap~(\%). MIS is evaluated on ER (700--800 nodes, 128 graphs) and RB (200--300 nodes, 500 graphs), reporting solved cost directly. All datasets follow prior work's standard splits: the TSP test sets follow DIFUSCO~\citep{sun2023difusco}, MIS-ER uses the DIMES evaluation graphs~\citep{qiu2022dimes}, and MIS-RB uses StruDiCO's evaluation graphs~\citep{wang2025strudico}.

\paragraph{PISG configuration.}
For TSP, three of the four AGNN solvers (DIFUSCO, T2T, FastT2T) share a single default: \texttt{adaptive}, \(s{=}2.0\), L7, \texttt{linear}. For MIS, per-distribution defaults are: ER uses \texttt{uniform}/\(s{=}2.0\)/L7/\texttt{constant}; RB uses \texttt{adaptive}/\(s{=}2.0\)/L7/\texttt{constant}.
On ER, these configurations transfer across three of four solvers; on RB, only FastT2T and StruDiCO have checkpoints. StruDiCO is a systematic exception: held-out validation selected substantially lower guidance strength on all tasks. TSP uses \texttt{adaptive}/\(s{=}0.1\)/L7/\texttt{linear}; MIS-ER uses \texttt{uniform}/\(s{=}0.1\)/L9/\texttt{constant}; MIS-RB uses \texttt{adaptive}/\(s{=}0.2\)/L5/\texttt{constant}. The calibration protocol and rationale appear in Appendix~\ref{app:strudico-calibration}.

\paragraph{Settings by subsection.}
Appendices~\ref{app:full-std} and~\ref{app:stat} use the controlled-isolation setting (\(T_g{=}0\), greedy decoding, no local search); the follow-up significance test in Appendix~\ref{app:stat} extends six cells to 20 paired seeds (41--60). Appendix~\ref{app:deployment} uses the deployment pipeline (\(T_g{>}0\) and 2-opt) with the main-table configurations frozen. Appendix~\ref{app:efficiency} measures runtime and memory under controlled isolation and compares PISG with baselines that use more denoising steps.

\subsection{Controlled-Isolation Results with Standard Deviations}
\label{app:full-std}

Tables~\ref{tab:tsp-small-std}--\ref{tab:mis-std} extend the main-text tables (Tables~\ref{tab:tsp}--\ref{tab:mis_main}) with per-seed standard deviations from 5 independent seeds (41--45). DIFUSCO and T2T report \(T_g{=}0\) only. In several cases PISG reduces cross-seed variance alongside improving the mean; for instance, DIFUSCO TSP-100 std drops from \(.067\) to \(.038\), and DIFUSCO MIS-ER std from \(.19\) to \(.09\).

\begin{table*}[t]
\caption{TSP small-scale results with standard deviations (gap\%, 5-seed mean\(\pm\)std). Extended version of Table~\ref{tab:tsp} (TSP-50/100). \textbf{Bold}: best gap within each solver block and scale.}
\label{tab:tsp-small-std}
\centering\scriptsize
\setlength{\tabcolsep}{2.0pt}
\begin{tabular*}{\textwidth}{@{\extracolsep{\fill}}ll ccc ccc@{}}
\toprule
& & \multicolumn{3}{c}{TSP-50} & \multicolumn{3}{c}{TSP-100} \\
\cmidrule(lr){3-5}\cmidrule(l){6-8}
Method & \(T_s/T_g\) & Cost & Gap\% & Time & Cost & Gap\% & Time \\
\midrule
Concorde & -- & 5.688 & 0.000 & -- & 7.756 & 0.000 & -- \\
\midrule
DIFUSCO & 100/0 & 5.708 & 0.367\(\pm\).044 & 6.2m & 7.864 & 1.388\(\pm\).067 & 6.7m \\
\;+PISG & 100/0 & 5.699 & \textbf{0.203}\(\pm\).017 & 6.2m & 7.800 & \textbf{0.567}\(\pm\).038 & 10.2m \\
\midrule
T2T & 50/0 & 5.727 & 0.683\(\pm\).071 & 3.5m & 7.882 & 1.624\(\pm\).064 & 3.6m \\
\;+PISG & 50/0 & 5.719 & \textbf{0.543}\(\pm\).034 & 3.5m & 7.815 & \textbf{0.758}\(\pm\).039 & 5.0m \\
\midrule
\multirow{2}{*}{FastT2T} & 5/0 & 5.689 & 0.023\(\pm\).005 & 45s & 7.762 & 0.079\(\pm\).007 & 52s \\
 & 5/5 & 5.688 & \textbf{0.0069}\(\pm\).0011 & 135s & 7.758 & 0.025\(\pm\).002 & 154s \\
\multirow{2}{*}{\;+PISG} & 5/0 & 5.689 & 0.017\(\pm\).001 & 48s & 7.760 & 0.051\(\pm\).003 & 55s \\
 & 5/5 & 5.688 & 0.0071\(\pm\).0012 & 141s & 7.758 & \textbf{0.023}\(\pm\).001 & 154s \\
\midrule
\multirow{2}{*}{StruDiCO} & 5/0 & 5.689 & 0.027\(\pm\).005 & 45s & 7.770 & 0.187\(\pm\).019 & 46s \\
 & 5/5 & 5.688 & 0.0125\(\pm\).001 & 71s & 7.758 & \textbf{0.031}\(\pm\).002 & 77s \\
\multirow{2}{*}{\;+PISG} & 5/0 & 5.689 & 0.025\(\pm\).003 & 48s & 7.770 & 0.179\(\pm\).015 & 53s \\
 & 5/5 & 5.688 & \textbf{0.0123}\(\pm\).001 & 70s & 7.758 & 0.031\(\pm\).002 & 82s \\
\bottomrule
\end{tabular*}
\end{table*}

\begin{table*}[t]
\caption{TSP large-scale results with standard deviations (gap\%, 5-seed mean\(\pm\)std). Extended version of Table~\ref{tab:tsp} (TSP-500/1000). \textbf{Bold}: best gap within each solver block and scale.}
\label{tab:tsp-large-std}
\centering\scriptsize
\setlength{\tabcolsep}{2.0pt}
\begin{tabular*}{\textwidth}{@{\extracolsep{\fill}}ll ccc ccc@{}}
\toprule
& & \multicolumn{3}{c}{TSP-500} & \multicolumn{3}{c}{TSP-1000} \\
\cmidrule(lr){3-5}\cmidrule(l){6-8}
Method & \(T_s/T_g\) & Cost & Gap\% & Time & Cost & Gap\% & Time \\
\midrule
Concorde & -- & 16.55 & 0.00 & -- & 23.12 & 0.00 & -- \\
\midrule
DIFUSCO & 100/0 & 18.52 & 11.91\(\pm\).22 & 3.8m & 26.19 & 13.31\(\pm\).29 & 9.6m \\
\;+PISG & 100/0 & 18.11 & \textbf{9.44}\(\pm\).38 & 5.8m & 25.65 & \textbf{10.95}\(\pm\).10 & 16.7m \\
\midrule
T2T & 50/0 & 18.42 & 11.33\(\pm\).42 & 76s & 26.13 & 13.05\(\pm\).12 & 2.4m \\
\;+PISG & 50/0 & 18.18 & \textbf{9.88}\(\pm\).45 & 96s & 26.09 & \textbf{12.87}\(\pm\).42 & 3.6m \\
\midrule
\multirow{2}{*}{FastT2T} & 5/0 & 17.51 & 5.84\(\pm\).41 & 32s & 24.61 & 6.46\(\pm\).20 & 48s \\
 & 5/5 & 16.92 & 2.27\(\pm\).13 & 68s & 23.99 & 3.78\(\pm\).05 & 177s \\
\multirow{2}{*}{\;+PISG} & 5/0 & 17.40 & 5.15\(\pm\).19 & 33s & 24.49 & 5.93\(\pm\).21 & 56s \\
 & 5/5 & 16.87 & \textbf{1.94}\(\pm\).09 & 69s & 23.86 & \textbf{3.22}\(\pm\).11 & 182s \\
\midrule
\multirow{2}{*}{StruDiCO} & 5/0 & 17.34 & 4.81\(\pm\).23 & 33s & 25.83 & 11.75\(\pm\).20 & 46s \\
 & 5/5 & 17.00 & 2.71\(\pm\).17 & 42s & 25.11 & 8.60\(\pm\).20 & 74s \\
\multirow{2}{*}{\;+PISG} & 5/0 & 17.33 & 4.74\(\pm\).26 & 37s & 25.43 & 9.99\(\pm\).15 & 58s \\
 & 5/5 & 16.95 & \textbf{2.45}\(\pm\).19 & 46s & 24.67 & \textbf{6.72}\(\pm\).34 & 86s \\
\bottomrule
\end{tabular*}
\end{table*}

\begin{table*}[t]
\caption{MIS results with standard deviations (cost\(\uparrow\), 5-seed mean\(\pm\)std). Extended version of Table~\ref{tab:mis_main}. \textbf{Bold}: best result within each solver block. \(^{*}\)Significant at \(\alpha{=}0.05\) (Table~\ref{tab:stat-significance}).}
\label{tab:mis-std}
\centering\scriptsize
\setlength{\tabcolsep}{1.5pt}
\begin{tabular*}{\textwidth}{@{\extracolsep{\fill}}ll ccc ccc@{}}
\toprule
& & \multicolumn{3}{c}{ER} & \multicolumn{3}{c}{RB} \\
\cmidrule(lr){3-5}\cmidrule(l){6-8}
Method & \(T_s/T_g\) & Cost\(\pm\)std\(\uparrow\) & Gap\%\(\downarrow\) & Time & Cost\(\pm\)std\(\uparrow\) & Gap\%\(\downarrow\) & Time \\
\midrule
Reference & -- & 41.35 & 0.00 & --- & 20.09 & 0.00 & --- \\
\midrule
DIFUSCO & 100/0 & 40.63\(\pm\).19 & 1.73 & 4.0m & --- & --- & --- \\
\;+PISG & 100/0 & 42.55\(\pm\).09 & \({-}\)\textbf{2.91}\(^{*}\) & 6.5m & --- & --- & --- \\
\midrule
T2T & 50/0 & 36.50\(\pm\).16 & 11.74 & 1.5m & --- & --- & --- \\
\;+PISG & 50/0 & 38.25\(\pm\).16 & \textbf{7.50}\(^{*}\) & 2.3m & --- & --- & --- \\
\midrule
\multirow{2}{*}{FastT2T} & 5/0 & 37.65\(\pm\).17 & 8.92 & 33s & 18.81\(\pm\).02 & 6.38 & 38s \\
 & 5/5 & 40.98\(\pm\).12 & 0.87 & 62s & 19.47\(\pm\).00 & 3.07 & 93s \\
\multirow{2}{*}{\;+PISG} & 5/0 & 38.89\(\pm\).30 & 5.92\(^{*}\) & 38s & 18.89\(\pm\).05 & 5.98\(^{*}\) & 41s \\
 & 5/5 & 41.86\(\pm\).07 & \({-}\)\textbf{1.25} & 67s & \textbf{19.49}\(\pm\).02 & \textbf{2.98} & 96s \\
\midrule
\multirow{2}{*}{StruDiCO} & 5/0 & 39.89\(\pm\).16 & 3.51 & 35s & 19.30\(\pm\).03 & 3.93 & 48s \\
 & 5/5 & 42.46\(\pm\).06 & \({-}\)2.72 & 49s & 19.74\(\pm\).02 & 1.77 & 73s \\
\multirow{2}{*}{\;+PISG} & 5/0 & 40.05\(\pm\).13 & 3.11 & 41s & 19.31\(\pm\).03 & 3.90 & 50s \\
 & 5/5 & 42.55\(\pm\).11 & \({-}\)\textbf{2.93} & 58s & 19.75\(\pm\).01 & \textbf{1.72} & 79s \\
\bottomrule
\end{tabular*}
\end{table*}

\subsection{Statistical Significance Analysis}
\label{app:stat}

Table~\ref{tab:stat-significance} reports paired statistical tests for each solver--task combination. For TSP (gap\%, \(\downarrow\)), we test \(H_1\): PISG reduces gap compared to baseline; for MIS (cost, \(\uparrow\)), we test \(H_1\): PISG increases cost. All tests use the exact one-sided Wilcoxon signed-rank test with \(n{=}5\) paired seeds (seeds 41--45). With \(n{=}5\), the minimum attainable \(p\)-value is \(0.031\); consequently, all \(p{<}0.05\) results in the table achieve this minimum.

Of the 22 comparisons, 16 reach significance at \(\alpha{=}0.05\) and 16 show 5/5 seed consistency. The six non-significant cases (T2T TSP-1000, StruDiCO TSP-50/100/500, StruDiCO MIS-ER, and StruDiCO MIS-RB) all show positive mean effects (PISG direction) over five seeds but with high variance relative to effect size. StruDiCO's lower guidance strength (\(s{=}0.1\)--\(0.2\)) produces smaller per-seed differences, requiring more seeds to resolve; the 20-seed follow-up below addresses these cells. Overall, PISG yields substantial, statistically significant gains on most settings, with marginal effects where the base solver is already near-optimal or the guidance strength is conservatively set. All reported $p$-values are uncorrected for multiple comparisons; conclusions rely on effect consistency across solvers, tasks, and scales rather than individual significance thresholds.

\begin{table*}[ht]
\caption{Statistical significance of PISG improvements. \textbf{Win}: seeds (out of 5) where PISG improves. \textbf{\(p\)}: one-sided Wilcoxon signed-rank. \textbf{\(\Delta\)}: mean improvement (\(>0\) = PISG better; TSP: baseline gap\% minus PISG gap\%; MIS: PISG cost minus baseline cost). Bold \(p\)-values: significant at \(\alpha{=}0.05\).}
\label{tab:stat-significance}
\centering
\scriptsize
\begin{tabular*}{\textwidth}{@{\extracolsep{\fill}}llccc@{}}
\toprule
Solver & Task & Win & \(p\) & \(\Delta\) (mean \(\pm\) std) \\
\midrule
DIFUSCO & TSP-50 & 5/5 & \textbf{0.031} & 0.1642 $\pm$ 0.0477 \\
DIFUSCO & TSP-100 & 5/5 & \textbf{0.031} & 0.8211 $\pm$ 0.0696 \\
DIFUSCO & TSP-500 & 5/5 & \textbf{0.031} & 2.4651 $\pm$ 0.2394 \\
DIFUSCO & TSP-1000 & 5/5 & \textbf{0.031} & 2.3601 $\pm$ 0.2495 \\
\midrule
T2T & TSP-50 & 5/5 & \textbf{0.031} & 0.1397 $\pm$ 0.0886 \\
T2T & TSP-100 & 5/5 & \textbf{0.031} & 0.8668 $\pm$ 0.0880 \\
T2T & TSP-500 & 5/5 & \textbf{0.031} & 1.4450 $\pm$ 0.7704 \\
T2T & TSP-1000 & 4/5 & 0.219 & 0.1803 $\pm$ 0.4110 \\
\midrule
FastT2T & TSP-50 & 5/5 & \textbf{0.031} & 0.0055 $\pm$ 0.0051 \\
FastT2T & TSP-100 & 5/5 & \textbf{0.031} & 0.0284 $\pm$ 0.0063 \\
FastT2T & TSP-500 & 5/5 & \textbf{0.031} & 0.6906 $\pm$ 0.3685 \\
FastT2T & TSP-1000 & 5/5 & \textbf{0.031} & 0.5264 $\pm$ 0.3562 \\
\midrule
StruDiCO & TSP-50 & 3/5 & 0.219 & 0.0014 $\pm$ 0.0034 \\
StruDiCO & TSP-100 & 4/5 & 0.312 & 0.0078 $\pm$ 0.0195 \\
StruDiCO & TSP-500 & 3/5 & 0.406 & 0.0677 $\pm$ 0.3971 \\
StruDiCO & TSP-1000 & 5/5 & \textbf{0.031} & 1.7573 $\pm$ 0.3436 \\
\midrule
\multicolumn{5}{l}{\textit{MIS}} \\
DIFUSCO & MIS-ER & 5/5 & \textbf{0.031} & 1.9172 $\pm$ 0.2664 \\
T2T & MIS-ER & 5/5 & \textbf{0.031} & 1.7500 $\pm$ 0.2196 \\
FastT2T & MIS-ER & 5/5 & \textbf{0.031} & 1.2391 $\pm$ 0.2892 \\
StruDiCO & MIS-ER & 4/5 & 0.156 & 0.1594 $\pm$ 0.2649 \\
\midrule
FastT2T & MIS-RB & 5/5 & \textbf{0.031} & 0.0776 $\pm$ 0.0390 \\
StruDiCO & MIS-RB & 2/5 & 0.406 & 0.0032 $\pm$ 0.0105 \\
\bottomrule
\end{tabular*}
\end{table*}

\paragraph{Follow-up with 20 paired seeds.}
Five seeds limit the attainable significance. We therefore extended the six non-significant cells to 20 paired seeds (41--60, including the original five) under the same controlled setting and the same one-sided Wilcoxon test (Table~\ref{tab:stat-20seed}). This is a follow-up test of these six cells, not a re-test of all 22 comparisons. Four cells reach \(p{<}0.05\): T2T TSP-1000 and StruDiCO TSP-100, TSP-500, and MIS-ER. All four remain significant after Holm correction across the six follow-up tests (largest adjusted \(p{=}0.036\)). StruDiCO TSP-50 remains inconclusive (\(p{=}0.062\), 13/20 wins) with a positive mean, and StruDiCO MIS-RB is effectively null (9/20 wins, mean \(\Delta\approx0\)). Taken together with Table~\ref{tab:stat-significance}, 15 of the 16 TSP solver--scale pairs and all four MIS-ER solvers reach \(p{<}0.05\) in either the original or the follow-up test.

\begin{table}[!ht]
\caption{Follow-up test of the six cells that were not significant with five seeds (controlled isolation, seeds 41--60, one-sided Wilcoxon signed-rank). \(\Delta\) as in Table~\ref{tab:stat-significance}, computed over 20 seeds. Bold: \(p{<}0.05\).}
\label{tab:stat-20seed}
\centering\scriptsize
\begin{tabular*}{\linewidth}{@{\extracolsep{\fill}}llcccc@{}}
\toprule
Solver & Task & Win (\(n{=}20\)) & \(p\) (\(n{=}20\)) & \(p\) (\(n{=}5\)) & \(\Delta\) (mean \(\pm\) std) \\
\midrule
T2T & TSP-1000 & 17/20 & \textbf{0.0032} & 0.219 & 0.2648 \(\pm\) 0.4110 \\
StruDiCO & TSP-50 & 13/20 & 0.0615 & 0.219 & 0.0030 \(\pm\) 0.0072 \\
StruDiCO & TSP-100 & 17/20 & \textbf{0.0008} & 0.312 & 0.0248 \(\pm\) 0.0310 \\
StruDiCO & TSP-500 & 14/20 & \textbf{0.0120} & 0.406 & 0.1417 \(\pm\) 0.2539 \\
StruDiCO & MIS-ER & 16/20 & \textbf{0.0047} & 0.156 & 0.1680 \(\pm\) 0.2265 \\
StruDiCO & MIS-RB & 9/20 & 0.5298 & 0.406 & \({-}\)0.0005 \(\pm\) 0.0288 \\
\bottomrule
\end{tabular*}
\end{table}

\subsection{Deployment Pipeline: Objective Rewrite and 2-opt}
\label{app:deployment}

The \(T_g{>}0\) rows of Tables~\ref{tab:tsp}--\ref{tab:mis_main} combine PISG with objective-side rewrite for FastT2T and StruDiCO but still omit local search. To test whether the pre-decoding benefit survives a complete deployment pipeline, we evaluate the two strongest solvers at the two largest TSP scales with \(T_s{=}5\), \(T_g{=}5\), and 2-opt applied after both the initial decode and the rewrite. 2-opt runs to convergence with a cap of 1{,}000 iterations; in a seed-42 diagnostic, both branches converged after about 30 iterations without reaching the cap. The main-table PISG configurations are frozen without re-tuning, and we use five paired seeds (41--45).

\begin{table}[ht]
\caption{Full deployment pipeline (\(T_s{=}5\), \(T_g{=}5\), 2-opt to convergence; seeds 41--45). Gap\% after the final rewrite and 2-opt. \(\Delta\): baseline gap minus PISG gap (\(>0\): PISG better).}
\label{tab:deployment}
\centering\scriptsize
\begin{tabular*}{\linewidth}{@{\extracolsep{\fill}}llcccc@{}}
\toprule
Solver & Task & Baseline & PISG & \(\Delta\) & Win \\
\midrule
FastT2T & TSP-500 & 0.429 & 0.403 & \(+\)0.026 & 4/5 \\
FastT2T & TSP-1000 & 0.627 & 0.599 & \(+\)0.028 & 4/5 \\
StruDiCO & TSP-500 & 0.323 & 0.314 & \(+\)0.009 & 4/5 \\
StruDiCO & TSP-1000 & 0.518 & 0.595 & \({-}\)0.076 & 0/5 \\
\bottomrule
\end{tabular*}
\end{table}

Table~\ref{tab:deployment} shows that PISG retains a positive mean effect in three of the four cells. On FastT2T, it removes \(6.1\%\) (TSP-500) and \(4.5\%\) (TSP-1000) of the gap remaining after rewrite and 2-opt, with 4/5 paired wins in each. StruDiCO TSP-500 shows a smaller positive effect (\(+0.009\) points, 4/5 wins). StruDiCO TSP-1000 reverses: PISG is worse than the baseline on all five seeds at the 2-opt endpoint. Gains are much smaller than under controlled isolation, which is consistent with downstream refinement absorbing part of the residual headroom.

\begin{table}[ht]
\caption{StruDiCO TSP-1000 after rewrite under increasing 2-opt budgets (seeds 41--45). The cap-0 row is the StruDiCO TSP-1000 \(T_g{=}5\) row of Table~\ref{tab:tsp}; the cap-1{,}000 row is the endpoint in Table~\ref{tab:deployment}.}
\label{tab:deployment-2opt}
\centering\scriptsize
\begin{tabular*}{\linewidth}{@{\extracolsep{\fill}}lcccc@{}}
\toprule
2-opt cap & Baseline & PISG & \(\Delta\) & Win \\
\midrule
0 & 8.60 & 6.72 & \(+\)1.88 & -- \\
10 & 2.048 & 1.678 & \(+\)0.370 & 5/5 \\
20 & 0.864 & 0.783 & \(+\)0.081 & 4/5 \\
1{,}000 (converged) & 0.518 & 0.595 & \({-}\)0.076 & 0/5 \\
\bottomrule
\end{tabular*}
\end{table}

Table~\ref{tab:deployment-2opt} traces the StruDiCO TSP-1000 reversal. The PISG advantage after rewrite persists through the first 20 2-opt iterations and reverses only at convergence. The frozen StruDiCO configuration therefore does not carry its pre-2-opt benefit to the converged endpoint at this scale.

\subsection{Compute Cost and Compute-Matched Comparison}
\label{app:efficiency}

\paragraph{Runtime and memory.}
PISG adds one perturbed forward pass per denoising step; the two branches are batched. Table~\ref{tab:efficiency} reports the overhead for both FastT2T and StruDiCO at \(T_s{=}5\), \(T_g{=}0\) across all six tasks. Wall-clock overhead is 3--17\% on FastT2T and 4--26\% on StruDiCO (5-seed mean). Memory overhead is negligible on TSP-50 (\(<\)1\%) and modest on TSP-100 (\(\approx\)8\%), rising to 27--112\% on larger-scale tasks where the second branch's intermediate activations dominate. StruDiCO's higher time overhead on TSP-1000 (\(+\)26\% vs.\ FastT2T's \(+\)17\%) reflects limited batching headroom in its objective-aware pipeline. In all cases the absolute overhead is small: the largest wall-clock increase is 12\,s (StruDiCO TSP-1000).

\begin{table*}[ht]
\caption{PISG overhead (FastT2T and StruDiCO, \(T_s{=}5\), \(T_g{=}0\), batched, 4\(\times\)A100-40GB). Batch size follows each solver's default; DIFUSCO/T2T overheads are proportional and can be read from Tables~\ref{tab:tsp}--\ref{tab:mis_main} time columns.}
\label{tab:efficiency}
\centering\scriptsize
\setlength{\tabcolsep}{3pt}
\begin{tabular*}{\textwidth}{@{\extracolsep{\fill}}l cccc cccc@{}}
\toprule
 & \multicolumn{4}{c}{FastT2T} & \multicolumn{4}{c}{StruDiCO} \\
\cmidrule(lr){2-5}\cmidrule(l){6-9}
 & \multicolumn{2}{c}{Time (s)} & \multicolumn{2}{c}{Mem (GB)} & \multicolumn{2}{c}{Time (s)} & \multicolumn{2}{c}{Mem (GB)} \\
\cmidrule(lr){2-3}\cmidrule(lr){4-5}\cmidrule(lr){6-7}\cmidrule(l){8-9}
Task & Base & +PISG & Base & +PISG & Base & +PISG & Base & +PISG \\
\midrule
TSP-50   & 45 & 48 & 0.71 & 0.71 & 45 & 48 & 0.71 & 0.71 \\
TSP-100  & 52 & 55 & 0.73 & 0.79 & 46 & 53 & 0.73 & 0.79 \\
TSP-500  & 32 & 33 & 0.92 & 1.32 & 33 & 37 & 0.93 & 1.32 \\
TSP-1000 & 48 & 56 & 1.57 & 3.10 & 46 & 58 & 1.57 & 3.10 \\
MIS-ER   & 33 & 38 & 2.06 & 4.14 & 35 & 41 & 1.85 & 3.93 \\
MIS-RB   & 38 & 41 & 0.88 & 1.48 & 48 & 50 & 1.21 & 1.54 \\
\bottomrule
\end{tabular*}
\end{table*}

\paragraph{Compute-matched comparison.}
\label{app:compute-equiv}
Table~\ref{tab:compute-equiv} asks whether the time PISG spends on the extra branch would be better spent on more denoising steps. For each task we compare PISG@\(T_s{=}5\) with (a)~the \emph{closest-time baseline}, the baseline \(T_s\) whose wall-clock time most closely matches PISG@\(T_s{=}5\), and (b)~the baseline at \(T_s{=}10\), which tests whether PISG can replace a doubled step budget. All six tasks are covered: TSP-100/500/1000 and MIS-ER come from the original dense \(T_s\) sweeps, and TSP-50 and MIS-RB from a five-seed quality rerun that reproduces the historical sweep values, with times taken from that sweep.

\begin{table}[ht]
\caption{Compute-matched comparison (FastT2T, \(T_g{=}0\), 5-seed mean). Closest-time: baseline \(T_s\) whose wall-clock \(\approx\) PISG@\(T_s{=}5\). Runtimes come from the dense \(T_s\) sweeps (independent measurements); absolute times can differ from Table~\ref{tab:efficiency}, so comparisons are made within each sweep. TSP: gap\% (\(\downarrow\)); MIS: solved cost (\(\uparrow\)).}
\label{tab:compute-equiv}
\centering\scriptsize
\begin{tabular*}{\textwidth}{@{\extracolsep{\fill}}lccccl@{}}
\toprule
Task & PISG@5 & Time & Closest base (\(T_s\)) & Base@10 & Outcome \\
\midrule
TSP-50 & 0.0171\% & 52\,s & \(T_s{=}5\): 0.0225\%, 52\,s & 0.0152\%, 71\,s & Beats closest-time; base@10 better \\
TSP-100 & 0.051\% & 54\,s & \(T_s{=}6\): 0.055\%, 56\,s & 0.042\%, 72\,s & Beats closest-time; base@10 better \\
TSP-500 & 5.15\% & 35\,s & \(T_s{=}8\): 5.48\%, 34\,s & 5.74\%, 38\,s & Beats both \\
TSP-1000 & 5.93\% & 58\,s & \(T_s{=}8\): 6.07\%, 57\,s & 6.15\%, 63\,s & Beats both \\
\midrule
MIS-ER & 38.89 & 37\,s & \(T_s{=}9\): 39.73, 37\,s & 39.79, 37\,s & Both baselines better \\
MIS-RB & 18.89 & 49\,s & \(T_s{=}6\): 18.92, 53\,s & 19.16, 68\,s & Both baselines better \\
\bottomrule
\end{tabular*}
\end{table}

On TSP, PISG@\(T_s{=}5\) outperforms the closest-time baseline at all four scales. Against the baseline at \(T_s{=}10\), it wins only on TSP-500 and TSP-1000, at comparable or lower wall-clock cost; on TSP-50 and TSP-100 the extra steps are better. On both MIS distributions, adding denoising steps is more effective than PISG at \(T_s{=}5\) under either comparison. The compute advantage of PISG is therefore limited to large-scale TSP in this comparison. PISG can still be combined with \(T_g\) scaling (Figure~\ref{fig:pareto}).

\clearpage
\section{Configuration Selection and Transfer}
\label{app:config-selection}

This appendix documents the three-level configuration strategy summarized in Section~\ref{sec:exp:sensitivity}: (1)~a full sensitivity sweep on FastT2T that identifies family defaults and a broad safe zone; (2)~direct transfer validation showing that these defaults work for DIFUSCO and T2T without re-tuning; (3)~a systematic held-out calibration for StruDiCO, whose redesigned noise type requires solver-specific parameters. Appendix~\ref{app:shift-transfer} then tests how frozen configurations transfer under density, scale, and real-world distribution shift.

PISG has four configuration axes per task--distribution pair:
\begin{itemize}
\item \textbf{Perturbation operator} $\mathcal{P}$: one of the three strategies defined in Appendix~\ref{app:perturbation-ops}.
\item \textbf{Guidance scale} $s$: global extrapolation strength. Family default \(s{=}2.0\); StruDiCO uses lower solver-specific values (\S\ref{app:strudico-calibration}).
\item \textbf{Layer set} $\mathcal{S}_{\text{pisg}}$: which backbone layers are perturbed; the shorthand ``L\(k\)'' perturbs a single layer at (0-indexed) position~\(k\). Default: L7 (layer index 7 in a 12-layer backbone).
\item \textbf{Schedule} $w(\cdot)$: temporal weighting (\texttt{constant} or \texttt{linear}); default varies by task.
\end{itemize}

\subsection{FastT2T: Full Sweep and Family Defaults}
\label{app:val-protocol}
\label{app:ablation}

\begin{figure}[t]
\centering
\includegraphics[width=0.6\linewidth]{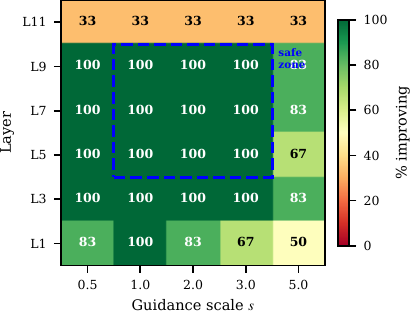}
\caption{TSP-100 sensitivity heatmap (FastT2T, seed~42). The main text (Section~\ref{sec:exp:sensitivity}) summarizes this sweep. 82\% of 180 gate-level configurations improve over the baseline; 100\% succeed in the safe zone (dashed box: scale~$\in[1,3]$, layers~$\in[5,9]$).}
\label{fig:sensitivity}
\end{figure}

The family defaults reported in Section~\ref{sec:exp:sensitivity} emerge from cross-task consistency patterns observed in a full sensitivity analysis on FastT2T. This analysis serves dual purposes: it maps the configuration landscape (Figure~\ref{fig:sensitivity}) and identifies robust default axes. A held-out validation search (below) selects the same default on MIS-ER and places the TSP default within the improving region on TSP-100.

\paragraph{Setup.}
We evaluate all six tasks (TSP-50/100/500/1000, MIS-ER, MIS-RB) with a factorial grid of 3~gate-level perturbation operators (uniform, adaptive, random-gate) \(\times\) 5~scales (\(s \in \{0.5, 1.0, 2.0, 3.0, 5.0\}\)) \(\times\) 6~layer positions (\(\text{L} \in \{1, 3, 5, 7, 9, 11\}\)) \(\times\) 2~schedules (constant, linear), yielding 180 configurations per task, plus a 10-configuration input-noise control (Appendix~\ref{app:perturbation-ops}) for a total of 190 configurations. All configurations are evaluated on the standard test set with seed~42.

\paragraph{Default selection.}
Table~\ref{tab:ablation-best} reports the per-task optimal configuration. Family defaults are chosen by favoring axes with the strongest cross-task consistency: Layer~7 is optimal on 4 of 6 subtasks; \(s{=}2.0\) provides a safe cross-task default (the per-task best scale is task-sensitive, but \(s{=}2.0\) always falls within the improving region); TSP tasks generally prefer \texttt{linear} schedules while MIS tasks prefer \texttt{constant}.

\paragraph{Per-task optima and safe zone.}

\begin{table*}[t]
\caption{Best FastT2T sweep configuration for each subtask. These are sweep optima, not necessarily the unified defaults used in the main tables (adap = adaptive, rand = random-gate, uni = uniform, const = constant).}
\label{tab:ablation-best}
\centering
\scriptsize
\begin{tabular*}{\textwidth}{@{\extracolsep{\fill}}llccc@{}}
\toprule
Subtask & Baseline & Best PISG & Rel. imp. & Best config \\
\midrule
TSP-50   & 0.0196 gap\% & 0.0143 & \(-27.3\%\) & adap / s1.0 / L5 / linear \\
TSP-100  & 0.0860 gap\% & 0.0414 & \(-51.9\%\) & rand / s3.0 / L7 / linear \\
TSP-500  & 5.6087 gap\% & 4.8040 & \(-14.3\%\) & uni / s3.0 / L7 / const \\
TSP-1000 & 6.3131 gap\% & 5.4532 & \(-13.6\%\) & adap / s1.0 / L7 / linear \\
MIS-ER   & 37.8125 & 39.2969 & \(+3.9\%\) & uni / s2.0 / L7 / const \\
MIS-RB   & 18.7740 & 19.1120 & \(+1.8\%\) & adap / s2.0 / L9 / const \\
\bottomrule
\end{tabular*}
\end{table*}

Three sweep-level patterns are robust.
\begin{itemize}
\item \textbf{Layer~7 is the best default perturbation location.} It gives the global optimum on 4 of 6 subtasks. Layer~11 yields the fewest improving configurations on every task (none on MIS-ER and TSP-50/500/1000); on TSP-500/1000, layer~9 also never improves and degrades more severely.
\item \textbf{TSP generally prefers linear schedules, but not without exception.} Three of four TSP subtasks favor \texttt{linear}; the exception is TSP-500, whose global optimum uses \texttt{constant}.
\item \textbf{Scale is task-sensitive rather than monotonic.} The safe default is \(s{=}2.0\) for both TSP and MIS, but the per-task best value is not a monotonic function of problem size.
\end{itemize}

\paragraph{Robustness across the configuration landscape.}
Of all 190 tested configurations (180 gate-level plus 10 input-noise), \(83\%\) improve over the baseline on TSP-100, \(68\%\) on MIS-RB, and \(35\%\) on TSP-50 (where the baseline is already near-optimal). Within the central region (scale~$\in [1, 3]$, layers~$\in [5, 9]$), all 54 configurations improve on TSP-100 but only 24 of 54 on TSP-500, where layer~9 degrades in every case; the safe range is therefore task-dependent.

\paragraph{Confirmatory held-out validation.}
To verify that the family defaults above are not artifacts of test-set selection, we repeat the configuration search on a held-out validation split. TSP-100 uses a 256/1024 (val/test) partition of the standard 1280-instance set; MIS-ER uses 32/96 out of 128 graphs. A 50-config grid (5~scales $\times$ 5~layer positions $\times$ 2~schedules, with perturbation family fixed per task; TSP minimizes gap\%, MIS maximizes solved cost) is swept on the val split with seed~42; the paper defaults are then evaluated on the test split with 5~seeds (41--45).

On MIS-ER, the validation optimum is the paper default. On TSP-100, the validation optimum is \texttt{adaptive}/\(s{=}3.0\)/L9/\texttt{constant}; the paper default ranks 15th of 50 and still improves over the baseline. The family default favors cross-task consistency over per-task optima (Default selection, above), and this optimum illustrates why: in the full sweep, layer~9 degrades every configuration on TSP-500 and TSP-1000. Applied unchanged to TSP-50, TSP-500, and TSP-1000, the TSP default consistently improves over the baseline without per-scale re-tuning (Table~\ref{tab:e13-test}).

\begin{table}[ht]
\caption{Held-out validation (5-seed mean$\pm$std on the test split). On MIS-ER the paper default coincides with the validation optimum. TSP-50/500/1000 rows apply the TSP default without re-tuning.}
\label{tab:e13-test}
\centering\scriptsize
\begin{tabular*}{\textwidth}{@{\extracolsep{\fill}}llcc@{}}
\toprule
Task & Config & Gap\% or Cost & Time (s) \\
\midrule
TSP-100 & Baseline & $0.081 \pm .016$ gap\% & 43 \\
TSP-100 & Paper default & $0.051 \pm .007$ gap\% & 46 \\
\midrule
MIS-ER & Baseline & $37.49 \pm 0.30$ cost & 28 \\
MIS-ER & Paper default (= val optimum) & $38.56 \pm 0.19$ cost & 32 \\
\midrule
TSP-50 & Baseline & $0.023 \pm .005$ gap\% & 44 \\
TSP-50 & Paper default (transfer) & $0.017 \pm .001$ gap\% & 46 \\
TSP-500 & Baseline & $5.84 \pm .41$ gap\% & 31 \\
TSP-500 & Paper default (transfer) & $5.15 \pm .19$ gap\% & 33 \\
TSP-1000 & Baseline & $6.46 \pm .20$ gap\% & 47 \\
TSP-1000 & Paper default (transfer) & $5.93 \pm .21$ gap\% & 56 \\
\bottomrule
\end{tabular*}
\end{table}

\subsection{DIFUSCO and T2T: Direct Transfer Validation}
\label{app:cross-solver}

The family defaults derived from FastT2T transfer directly to DIFUSCO and T2T without re-tuning. Tables~\ref{tab:cross-solver-tsp} and~\ref{tab:cross-solver-mis} verify this by applying the shared default and single-axis variants to each solver (seed~42).

\begin{table}[ht]
\caption{TSP-100 cross-solver transfer (gap\%, \(\downarrow\), seed\,42). The shared default from FastT2T (adaptive/\(s{=}2.0\)/L7/linear) is applied to DIFUSCO and T2T. Each row varies one axis; bold indicates improvement over baseline. Scale variants: \(\downarrow{=}1.0\), \(\uparrow{=}3.0\).}
\label{tab:cross-solver-tsp}
\centering
\scriptsize
\begin{tabular*}{\textwidth}{@{\extracolsep{\fill}}lcc@{}}
\toprule
& DIFUSCO & T2T \\
\midrule
Baseline & 1.393 & 1.559 \\
Default & \textbf{0.561} & \textbf{0.812} \\
\midrule
Layer \(\to\) 5 & 1.698 & 1.665 \\
Layer \(\to\) 9 & \textbf{0.845} & \textbf{1.437} \\
Scale \(\downarrow\) & \textbf{0.698} & \textbf{0.978} \\
Scale \(\uparrow\) & \textbf{0.472} & \textbf{0.741} \\
\(\to\) constant & \textbf{0.483} & \textbf{0.805} \\
\(\to\) uniform & \textbf{0.846} & \textbf{1.084} \\
\(\to\) random-gate & \textbf{0.728} & \textbf{0.962} \\
\bottomrule
\end{tabular*}
\end{table}

\begin{table}[ht]
\caption{MIS-ER cross-solver transfer (cost, \(\uparrow\), seed\,42). The shared default from FastT2T (uniform/\(s{=}2.0\)/L7/constant) is applied to DIFUSCO and T2T. Each row varies one axis; bold indicates improvement over baseline. Scale variants: \(\downarrow{=}1.0\), \(\uparrow{=}3.0\).}
\label{tab:cross-solver-mis}
\centering
\scriptsize
\begin{tabular*}{\textwidth}{@{\extracolsep{\fill}}lcc@{}}
\toprule
& DIFUSCO & T2T \\
\midrule
Baseline & 40.38 & 36.76 \\
Default & \textbf{42.59} & \textbf{38.20} \\
\midrule
Layer \(\to\) 5 & \textbf{41.56} & \textbf{38.80} \\
Layer \(\to\) 9 & 39.78 & \textbf{39.98} \\
Scale \(\downarrow\) & \textbf{42.00} & \textbf{38.24} \\
Scale \(\uparrow\) & \textbf{42.86} & \textbf{38.47} \\
\(\to\) linear & \textbf{41.91} & \textbf{38.14} \\
\(\to\) adaptive & \textbf{41.84} & \textbf{41.63} \\
\(\to\) random-gate & \textbf{42.59} & \textbf{38.50} \\
\bottomrule
\end{tabular*}
\end{table}

\paragraph{Transfer patterns.}
On TSP-100, the shared default improves both DIFUSCO and T2T; 14 of 16 solver--configuration cells (88\%) improve over the respective baseline, the only exceptions being Layer~5. On MIS-ER, 15 of 16 cells improve, the exception being DIFUSCO with Layer~9. The shared default (Layer~7) improves every cell on both tasks, consistent with the single-solver landscape analysis.

On MIS-ER, \texttt{uniform} remains the family default rather than \texttt{adaptive} because \texttt{adaptive} severely degrades FastT2T on MIS-ER (Table~\ref{tab:perturbation}), making it an unreliable cross-solver choice.

\subsection{StruDiCO: Solver-Specific Calibration}
\label{app:strudico-calibration}

StruDiCO employs variable-absorption noising and a redesigned edge-gating architecture, making it qualitatively different from the other three solvers in the family. The shared family default (\(s{=}2.0\)) severely degrades StruDiCO on MIS-ER, and even moderate scales (\(s{\geq}1.0\)) cause instability on TSP (Table~\ref{tab:strudico-stage1}). We therefore perform a systematic coarse-to-fine held-out calibration to identify solver-specific configurations.

\paragraph{Stage~1: Coarse transfer check.}
We evaluate the family default scale and three alternatives (\(s \in \{0.5, 1.0, 2.0, 3.0\}\)) on held-out validation splits for TSP-100, MIS-ER, and MIS-RB (3~seeds each). Table~\ref{tab:strudico-stage1} shows that higher scales monotonically degrade all three tasks; \(s{=}0.5\) is the best coarse candidate but still underperforms the baseline on MIS, motivating a finer search in the low-scale region.

\begin{table}[ht]
\caption{StruDiCO coarse transfer check (Stage~1, 3-seed mean on validation splits). Family default scale (\(s{=}2.0\)) degrades all three tasks. Arrows indicate metric direction (\(\downarrow\)~gap\%, \(\uparrow\)~cost).}
\label{tab:strudico-stage1}
\centering
\scriptsize
\begin{tabular*}{\textwidth}{@{\extracolsep{\fill}}lccc@{}}
\toprule
Scale & TSP-100 gap\% \((\downarrow)\) & MIS-ER cost \((\uparrow)\) & MIS-RB cost \((\uparrow)\) \\
\midrule
Baseline & 0.172 & 39.85 & 19.30 \\
\(s{=}0.5\) & \textbf{0.178} & 39.20 & 19.30 \\
\(s{=}1.0\) & 0.248 & 38.73 & 19.14 \\
\(s{=}2.0\) & 0.518 & 34.90 & 18.78 \\
\(s{=}3.0\) & 2.448 & 33.35 & 18.68 \\
\bottomrule
\end{tabular*}
\end{table}

\paragraph{Stages~2--3: Low-scale refinement and single-axis sweep.}
Stage~2 explores the low-scale region (\(s \in \{0.1, 0.2, 0.3, 0.7\}\)) and identifies per-task scale optima: \(s{=}0.1\) for TSP-100 and MIS-ER, \(s{=}0.2\) for MIS-RB. Stage~3 fixes the selected scale and varies one axis at a time to check whether departing from the transferred defaults yields improvement. The axes tested are layer (\{5, 7, 9\}), perturbation operator, schedule (\{constant, linear, reverse-linear\}), fusion space (\{logit, probability\}; probability fusion was explored but not adopted in any final configuration), and active steps (\{all, last\}). For TSP-100, the transferred default (adaptive/L7/linear) remains optimal; for MIS-RB, calibration selected L5 over the transferred L7 (Table~\ref{tab:strudico-frozen}); for MIS-ER, Layer~9 improves over Layer~7 and is adopted. A small composed-candidate check (Stage~3b) verifies that the per-axis winners compose safely; for MIS-ER, the single-axis Layer~9 winner outperforms multi-axis combinations.

\paragraph{Frozen configurations.}
Table~\ref{tab:strudico-frozen} summarizes the calibrated configurations used in all main-table StruDiCO results. These are frozen before test evaluation; the held-out test split is accessed only once with the final configuration.

\begin{table}[ht]
\caption{StruDiCO frozen configurations from held-out calibration. ``Transferred default'' shows the family default axis value; boldface marks axes where calibration selected a different value.}
\label{tab:strudico-frozen}
\centering
\scriptsize
\begin{tabular*}{\textwidth}{@{\extracolsep{\fill}}lccc@{}}
\toprule
Axis & TSP & MIS-ER & MIS-RB \\
\midrule
Perturbation & adaptive & uniform & adaptive \\
Scale (\(s\)) & \textbf{0.1} & \textbf{0.1} & \textbf{0.2} \\
Layers & L7 & \textbf{L9} & \textbf{L5} \\
Schedule & linear & constant & constant \\
Fusion & logit & logit & logit \\
\midrule
\multicolumn{4}{l}{\textit{Transferred default: adaptive (TSP) / uniform (MIS-ER) / adaptive (MIS-RB) / \(s{=}2.0\) / L7 / task-default schedule / logit}} \\
\bottomrule
\end{tabular*}
\end{table}

\paragraph{Calibration cost.}
The entire calibration pipeline evaluates 54 configurations \(\times\) 3~seeds \(=\) 162 runs and completes in \({\sim}\)80\,minutes on 4~GPUs. This modest cost is comparable to one full 5-seed test evaluation and is performed once per solver.

\subsection{Frozen-Configuration Transfer under Distribution Shift}
\label{app:shift-transfer}

The preceding subsections select configurations for each task--distribution pair. This subsection evaluates how a selected configuration behaves when it is frozen and the test distribution moves away from the one it was selected on. We probe three axes separately: graph density, problem scale, and real-world TSP instances. All runs use controlled isolation (\(T_g{=}0\), greedy decoding, no local search) and paired seeds 41--45.

\paragraph{Density.}
We evaluate the FastT2T MIS-ER checkpoint, trained at edge probability \(p{=}0.15\), on ER graphs with \(n{=}750\) and \(p\in\{0.10,0.125,0.15,0.175,0.20\}\) (128 graphs per density), using the MIS-ER family default (\texttt{uniform}/\(s{=}2.0\)) for all densities. Table~\ref{tab:shift-density} reports relative gains with 95\% intervals from a crossed paired graph-and-seed bootstrap. Transfer is positive from \(p{=}0.10\) to \(p{=}0.175\), with all intervals excluding zero and 5/5 wins, and negative at \(p{=}0.20\). The boundary is asymmetric: the shift to sparser graphs (\(p{=}0.10\)) succeeds, while the shift of equal size to denser graphs (\(p{=}0.20\)) fails.

\begin{table}[ht]
\caption{Density transfer on MIS-ER (FastT2T trained at \(p{=}0.15\), \(n{=}750\), frozen \texttt{uniform}/\(s{=}2.0\)). Relative gain \(=100\times(\text{PISG}-\text{baseline})/\text{baseline}\).}
\label{tab:shift-density}
\centering\scriptsize
\begin{tabular*}{\linewidth}{@{\extracolsep{\fill}}ccccc@{}}
\toprule
\(p\) & Baseline & PISG & Relative gain (95\% CI) & Win \\
\midrule
0.100 & 50.85 & 52.17 & \(+\)2.67\% [0.98, 4.38] & 5/5 \\
0.125 & 44.84 & 46.50 & \(+\)3.93\% [2.99, 4.88] & 5/5 \\
0.150 & 37.59 & 38.64 & \(+\)3.08\% [1.90, 4.33] & 5/5 \\
0.175 & 31.47 & 32.03 & \(+\)2.13\% [0.86, 3.38] & 5/5 \\
0.200 & 25.35 & 24.69 & \({-}\)2.28\% [\({-}\)3.49, \({-}\)0.99] & 0/5 \\
\bottomrule
\end{tabular*}
\end{table}

\paragraph{Scale.}
We freeze the TSP-100-selected setting (\(s{=}2.0\), \texttt{linear} schedule) and evaluate FastT2T checkpoints trained at four scales on all four test scales (Table~\ref{tab:shift-scale}). When the test scale does not exceed the training scale, the frozen configuration improves 9 of 10 cells; the exception is the TSP-1000 checkpoint tested at TSP-500. Every cell is unanimous across seeds (5/5 or 0/5). All other negative cells test a checkpoint on instances larger than those it was trained on.

\begin{table}[ht]
\caption{Scale transfer on FastT2T with a frozen TSP-100-selected configuration. Entries are mean gap reductions in percentage points (\(>0\): PISG better) with paired wins. \(^\dagger\): test scale larger than training scale.}
\label{tab:shift-scale}
\centering\scriptsize
\begin{tabular*}{\linewidth}{@{\extracolsep{\fill}}lcccc@{}}
\toprule
Train \(\backslash\) Test & 50 & 100 & 500 & 1000 \\
\midrule
50 & \(+\)0.01 (5/5) & \(+\)0.43 (5/5)\(^\dagger\) & \({-}\)7.18 (0/5)\(^\dagger\) & \({-}\)7.45 (0/5)\(^\dagger\) \\
100 & \(+\)0.01 (5/5) & \(+\)0.03 (5/5) & \({-}\)6.76 (0/5)\(^\dagger\) & \({-}\)2.78 (0/5)\(^\dagger\) \\
500 & \(+\)0.89 (5/5) & \(+\)1.38 (5/5) & \(+\)0.69 (5/5) & \(+\)0.94 (5/5)\(^\dagger\) \\
1000 & \(+\)0.71 (5/5) & \(+\)0.62 (5/5) & \({-}\)0.87 (0/5) & \(+\)0.53 (5/5) \\
\bottomrule
\end{tabular*}
\end{table}

\paragraph{TSPLIB.}
TSPLIB~\citep{reinelt1991tsplib} instances combine a non-uniform spatial distribution with a scale mismatch. We evaluate DIFUSCO, T2T, and FastT2T with their frozen TSP defaults on two bands: 50--200 nodes (29 instances, TSP-100 checkpoints) and 200--1{,}000 nodes (19 instances, TSP-500 checkpoints). Table~\ref{tab:shift-tsplib} shows that in the 50--200 band, DIFUSCO and FastT2T improve with 4/5 wins each, while T2T degrades. In the 200--1{,}000 band, all three solvers are neutral or negative. Because this shift changes several factors at once, it does not isolate which factor causes the failures.

\begin{table}[ht]
\caption{TSPLIB transfer with frozen TSP defaults (controlled isolation, seeds 41--45). Gap\% and mean gap reduction \(\Delta\) in percentage points (\(>0\): PISG better).}
\label{tab:shift-tsplib}
\centering\scriptsize
\begin{tabular*}{\linewidth}{@{\extracolsep{\fill}}llcccc@{}}
\toprule
Band & Solver & Baseline & PISG & \(\Delta\) & Win \\
\midrule
50--200 & DIFUSCO & 4.51 & 4.05 & \(+\)0.46 & 4/5 \\
50--200 & T2T & 4.66 & 5.12 & \({-}\)0.46 & 2/5 \\
50--200 & FastT2T & 1.73 & 1.68 & \(+\)0.05 & 4/5 \\
\midrule
200--1{,}000 & DIFUSCO & 19.77 & 19.91 & \({-}\)0.14 & 3/5 \\
200--1{,}000 & T2T & 38.77 & 40.62 & \({-}\)1.85 & 1/5 \\
200--1{,}000 & FastT2T & 30.87 & 32.11 & \({-}\)1.24 & 2/5 \\
\bottomrule
\end{tabular*}
\end{table}

\paragraph{Summary.}
A frozen configuration transfers within the coverage of its training distribution: across moderate density changes, to test scales no larger than the training scale, and partly to small real-world instances. Failures concentrate at the high-density endpoint, at most extrapolations to larger scales, and at large TSPLIB instances. These results delimit the claim in Section~\ref{sec:exp:sensitivity} that good configuration regions transfer within a task--distribution pair.

\clearpage
\section{Additional Mechanism Evidence}
\label{app:mechanism}

This appendix provides additional evidence for the signal characterization claims in Section~\ref{sec:signal-characterization}: the signal concentrates on structurally contested decisions (not isotropic rescaling), corrections compose along the trajectory (not step-by-step improvement), and these signatures are consistent across the edge-gated AGNN family. Appendices~\ref{app:diagnostics-quality} and~\ref{app:subtour} delimit what the diagnostics show: their association with final quality, and the absence of subtour-level signal in TSP.

\subsection{Raw-output Constraint-Boundary Alignment}
\label{app:raw-output-alignment}

\paragraph{Purpose.}
This experiment tests whether the trained denoiser's raw output---before any greedy decoding or post-processing---already exhibits constraint-aligned responses, and whether PISG amplifies them.

\paragraph{Three branches.}
We run three \emph{independent}, complete denoising trajectories from the same trained FastT2T checkpoint and the same initial noise \(\mathbf{x}_{T_s}\), differing only in the PISG configuration:
\begin{itemize}
\item \textbf{Normal}: standard inference without perturbation or guidance.
\item \textbf{Perturbed}: structure-degraded gating is applied at every step, but \emph{no} extrapolation is performed; the model's raw prediction under weakened gating is used directly.
\item \textbf{Guided}: full PISG inference (Algorithm~\ref{alg:pisg}).
\end{itemize}
Each branch produces an independent set of raw edge/node probabilities, which are then passed to the standard greedy decoder for evaluation.

\paragraph{Metrics.}
For TSP, we measure (i)~the optimality gap~(\%) to the Concorde reference, and (ii)~the total number of edges the greedy decoder skips due to degree-2 violations during tour construction:
\begin{equation}
\text{TotalSkips} = \sum_{i=1}^{n} \bigl|\{e \in \text{candidates}(i) : e \text{ skipped due to degree-2 violation}\}\bigr|,
\end{equation}
where \(\text{candidates}(i)\) are the edges considered at step \(i\) of greedy tour construction. For MIS, we measure (i)~the independent-set size and (ii)~the probability separation between selected and excluded nodes:
\begin{equation}
\text{ProbSep} = \frac{1}{|S|}\sum_{i \in S} p_i \;-\; \frac{1}{|V \setminus S|}\sum_{j \notin S} p_j,
\end{equation}
where \(S\) is the set of nodes selected by the greedy decoder and \(p_i\) is the raw predicted probability for node~\(i\).

\paragraph{Procedure.}
We use a trained FastT2T checkpoint on 128 instances per TSP scale (TSP-50/100/500/1000) and 128 graphs for MIS-ER (seed~42). All three branches (normal, guided, perturbed) share the same initial noise per instance to ensure a fair comparison; the noise seed is reset per instance, so absolute gap values are not directly comparable to the main-table results (which use independent noise across instances and 5-seed aggregation). The relative ordering among branches remains valid. Tables~\ref{tab:e22-tsp} and~\ref{tab:e22-mis} report the underlying data for Figure~\ref{fig:motivation}(b).

\noindent\textit{Note:} Figure~\ref{fig:motivation}(a) uses a 10-node TSP instance constructed to illustrate the degree-2 constraint pattern; it is a schematic diagram, not the output of an actual solver.

\begin{table}[h]
\caption{TSP raw-output metrics underlying Figure~\ref{fig:motivation}(b) (128 instances per scale, single seed).}
\label{tab:e22-tsp}
\centering\scriptsize
\begin{tabular*}{\textwidth}{@{\extracolsep{\fill}}llrr@{}}
\toprule
Scale & Branch & gap (\%) & total skips \\
\midrule
\multirow{3}{*}{TSP-50}   & Perturbed & 1.586 & 32.4 \\
                           & Normal    & 0.127 & 4.6 \\
                           & Guided    & 0.040 & 0.08 \\
\addlinespace
\multirow{3}{*}{TSP-100}  & Perturbed & 5.711 & 788.4 \\
                           & Normal    & 0.339 & 43.3 \\
                           & Guided    & 0.239 & 30.9 \\
\addlinespace
\multirow{3}{*}{TSP-500}  & Perturbed & 15.308 & 68277 \\
                           & Normal    & 6.103 & 47437 \\
                           & Guided    & 5.434 & 34258 \\
\addlinespace
\multirow{3}{*}{TSP-1000} & Perturbed & 16.364 & 279783 \\
                           & Normal    & 8.927 & 240013 \\
                           & Guided    & 5.153 & 149813 \\
\bottomrule
\end{tabular*}
\end{table}

\begin{table}[h]
\caption{MIS-ER raw-output metrics underlying Figure~\ref{fig:motivation}(b) (128 graphs, single seed).}
\label{tab:e22-mis}
\centering\scriptsize
\begin{tabular*}{\textwidth}{@{\extracolsep{\fill}}lrr@{}}
\toprule
Branch & IS size & prob separation \\
\midrule
Perturbed & 34.52 & 0.448 \\
Normal    & 37.54 & 0.787 \\
Guided    & 39.49 & 0.886 \\
\bottomrule
\end{tabular*}
\end{table}

\subsection{TSP: Degree-2 Sign Compliance Details}
\label{app:tsp-signal-detail}

\paragraph{Purpose.}
This experiment tests whether the probability-space guidance signal \(\boldsymbol{\delta} = \psi(\mathbf{p}^{\text{normal}}) - \psi(\mathbf{p}^{\text{perturbed}})\) (denoted \(\tilde{\boldsymbol{\delta}}_t\) in Section~\ref{sec:signal-characterization}; \(\boldsymbol{\delta}\) in this subsection and Appendix~\ref{app:mis-signal-detail}) exhibits a non-random sign structure aligned with the TSP degree-2 tour constraint, rather than acting as isotropic rescaling.

\paragraph{Metric definition.}
\emph{Sign Compliance at degree~2} (sc@2) measures, for each node \(i\), whether \(\boldsymbol{\delta}\) simultaneously boosts the top-2 candidate edges and suppresses the third-ranked edge:
\begin{equation}
\text{sc@2}(i) = \mathbb{1}[\delta_{e_0(i)} > 0] \cdot \mathbb{1}[\delta_{e_1(i)} > 0] \cdot \mathbb{1}[\delta_{e_2(i)} < 0],
\end{equation}
where \(e_k(i)\) denotes the edge ranked \(k\)-th by normal-branch probability among all edges incident to node \(i\) (rank~0 = highest). The global sc@2 is the fraction of nodes satisfying this joint condition. As a null baseline, we compute a \emph{shuffled} sc@2 by randomly permuting \(\boldsymbol{\delta}\) across edges within each instance and averaging over 100 permutations.

\paragraph{Procedure.}
We compute sc@2 at each of the \(T_s{=}5\) denoising steps on TSP-100 using FastT2T over five seeds (41--45), with 1280 instances per seed and 6400 instance-seed pairs in total. Statistical significance is assessed via 100-permutation tests, one-sided Wilcoxon signed-rank tests, and bootstrap 95\% confidence intervals (1000 resamples).

\paragraph{Results.}

\begin{table}[t]
\caption{sc@2 summary across denoising steps (TSP-100, 5 seeds, 6400 instance-seed pairs). Ratio = sc@2 / shuffled baseline. All permutation $p < 0.01$ and Wilcoxon $p < 10^{-95}$.}
\label{tab:sc2-summary}
\centering
\scriptsize
\begin{tabular*}{\textwidth}{@{\extracolsep{\fill}}ccccccc@{}}
\toprule
Step & $t$ & sc@2 & Shuffled & Ratio & sc@2 95\% CI & Ratio 95\% CI \\
\midrule
0 & 1000 & 0.764 & 0.041 & $18.6\times$ & [0.761, 0.767] & [18.3, 18.9] \\
1 & 691  & 0.764 & 0.041 & $18.6\times$ & [0.761, 0.768] & [18.3, 18.9] \\
2 & 413  & \textbf{0.782} & 0.036 & $\mathbf{22.0\times}$ & [0.779, 0.785] & [21.7, 22.4] \\
3 & 191  & 0.425 & 0.020 & $21.1\times$ & [0.420, 0.429] & [20.6, 21.5] \\
4 & 49   & 0.021 & 0.007 & $3.0\times$  & [0.021, 0.022] & [2.9, 3.2]   \\
\bottomrule
\end{tabular*}
\end{table}

\begin{table}[t]
\caption{Three-condition decomposition of sc@2 (TSP-100). Each column shows the fraction of nodes satisfying that individual condition. sc@2 is the fraction satisfying all three conditions jointly.}
\label{tab:sc2-decomposition}
\centering
\scriptsize
\begin{tabular*}{\textwidth}{@{\extracolsep{\fill}}cccccc@{}}
\toprule
Step & $\delta_{\text{rank}\,0}{>}0$ & $\delta_{\text{rank}\,1}{>}0$ & $\delta_{\text{rank}\,2}{<}0$ & $\delta_{\text{rest}}{<}0$ & sc@2 \\
\midrule
0 & 91.8\% & 90.5\% & 85.5\% & 78.9\% & 76.4\% \\
1 & 91.8\% & 90.5\% & 85.5\% & 78.9\% & 76.4\% \\
2 & 92.0\% & 92.0\% & 86.8\% & 81.2\% & 78.2\% \\
3 & 59.0\% & 72.9\% & 86.5\% & 86.5\% & 42.5\% \\
4 &  9.8\% & 11.2\% & 85.1\% & 92.1\% &  2.1\% \\
\bottomrule
\end{tabular*}
\end{table}

Table~\ref{tab:sc2-summary} shows that sc@2 is $76$--$78\%$ at steps~0--2 and decays sharply afterward. Table~\ref{tab:sc2-decomposition} decomposes sc@2 into its three constituent conditions: at steps~0--2, all three conditions exceed $85\%$. At steps~3--4, the positive conditions for rank~0 and rank~1 drop sharply (reflecting prediction convergence), while the rank-2 negative condition remains stable at ${\geq}85\%$, indicating that suppression of the third-ranked edge is the most persistent component of the degree-2 signal.

\begin{table}[t]
\caption{Direction alignment on matched-subset TSP-100 trajectories (5 seeds, 6400 instance-seed pairs). Preference ratio is \(\delta_{\text{opt}} / (|\delta_{\text{opt}}| + |\delta_{\text{nonopt}}|)\) averaged across instances.}
\label{tab:branch-alignment}
\centering
\scriptsize
\begin{tabular*}{\textwidth}{@{\extracolsep{\fill}}ccccc@{}}
\toprule
Step & Pref. ratio & 95\% CI & \(|\delta_{\text{opt}}|{>}|\delta_{\text{nonopt}}|\) frac. & opt.\ positive frac. \\
\midrule
0 & 0.884 & [0.873, 0.893] & 99.97\% & 89.8\% \\
1 & 0.883 & [0.874, 0.892] & 99.95\% & 89.8\% \\
2 & 0.862 & [0.851, 0.873] & 99.92\% & 90.2\% \\
3 & 0.501 & [0.482, 0.521] & 99.50\% & 63.1\% \\
4 & 0.524 & [0.506, 0.542] & 94.12\% & 8.8\% \\
\bottomrule
\end{tabular*}
\end{table}

As a complementary view, Table~\ref{tab:branch-alignment} decomposes \(\boldsymbol{\delta}\) by optimal-edge membership. At steps~0--2, the preference ratio is \(0.86\)--\(0.88\), the optimal-edge direction is positive for about \(90\%\) of instances, and \(|\delta_{\text{opt}}|\) exceeds \(|\delta_{\text{nonopt}}|\) in more than \(99.9\%\) of instances, consistent with the sc@2 sign pattern.

\subsection{MIS: Conflict-Edge Discrimination Details}
\label{app:mis-signal-detail}

\paragraph{Purpose.}
This experiment tests whether the probability-space signal \(\boldsymbol{\delta}\) (Appendix~\ref{app:tsp-signal-detail}) encodes constraint-relevant magnitude differences on conflict edges in MIS. Unlike TSP, MIS \(\boldsymbol{\delta}\) is predominantly positive (\(>99\%\) for high-probability nodes); the constraint signal is expected to appear as magnitude asymmetry rather than sign reversal.

\paragraph{Metric definition.}
\emph{Conflict Magnitude Asymmetry} (CMA) quantifies how much \(\boldsymbol{\delta}\) differentiates the two endpoints of an edge. Given a probability threshold \(\theta\), an edge \((i,j) \in E\) is a \emph{conflict edge} if both endpoints have high predicted selection probability: \(\sigma(p_i^{\text{normal}}) > \theta\) and \(\sigma(p_j^{\text{normal}}) > \theta\). CMA on a set of edges \(\mathcal{C}\) is defined as:
\begin{equation}
\text{CMA}_{\mathcal{C}} = \frac{1}{|\mathcal{C}|}\sum_{(i,j) \in \mathcal{C}} |\delta_i - \delta_j|.
\end{equation}
The CMA ratio compares conflict edges to non-conflict edges: \(\text{Ratio} = \text{CMA}_{\text{conflict}} / \text{CMA}_{\text{non-conflict}}\). A shuffled baseline randomly permutes \(\boldsymbol{\delta}\) across nodes within each instance (100 permutations).

\paragraph{Procedure.}
We compute CMA at each denoising step on MIS-ER using FastT2T over five seeds (41--45), with 128 graphs per seed and 640 instance-seed pairs in total. We evaluate thresholds \(\theta \in \{0.3, 0.5, 0.7\}\). Statistical significance is assessed via permutation tests, one-sided Wilcoxon signed-rank tests (per-instance \(\text{CMA}_{\text{conflict}}\) vs.\ \(\text{CMA}_{\text{non-conflict}}\)), and bootstrap 95\% CIs (1000 resamples).

\paragraph{Results.}

\begin{table}[t]
\caption{CMA across denoising steps and thresholds (MIS-ER, 5 seeds, 640 instance-seed pairs). CMA = mean $|\delta_i - \delta_j|$. Ratio = conflict / non-conflict. Rows for $\theta{=}0.5$ and $0.7$ at steps 0--2 are omitted because conflict edges occur in at most 10 of the 640 instance-seed pairs.}
\label{tab:cma-detail}
\centering
\scriptsize
\begin{tabular*}{\textwidth}{@{\extracolsep{\fill}}ccccccc@{}}
\toprule
Step & $\theta$ & $n_{\text{conflict}}$ & CMA$_{\text{conflict}}$ & CMA$_{\text{non-conf}}$ & CMA$_{\text{shuffled}}$ & Ratio \\
\midrule
0 & 0.3 &   1823 & 0.056 & 0.037 & 0.037 & $1.5\times$ \\
1 & 0.3 &   1883 & 0.056 & 0.038 & 0.037 & $1.5\times$ \\
2 & 0.3 &   1934 & 0.086 & 0.029 & 0.029 & $2.9\times$ \\
3 & 0.3 &  50639 & 0.189 & 0.070 & 0.070 & $2.7\times$ \\
3 & 0.5 &   7382 & 0.200 & 0.072 & 0.072 & $2.8\times$ \\
3 & 0.7 &    548 & 0.249 & 0.076 & 0.076 & $3.3\times$ \\
4 & 0.3 &  39861 & \textbf{0.233} & 0.055 & 0.056 & $\mathbf{4.2\times}$ \\
4 & 0.5 &   9981 & \textbf{0.232} & 0.057 & 0.057 & $4.1\times$ \\
4 & 0.7 &   1449 & \textbf{0.241} & 0.060 & 0.060 & $4.0\times$ \\
\bottomrule
\end{tabular*}
\end{table}

\begin{table}[t]
\caption{Hypothesis tests for CMA ratio (MIS-ER, 5-seed aggregate). Permutation: 100 shuffles of $\delta$ across nodes; Wilcoxon: one-sided paired test on per-instance CMA$_{\text{conflict}}$ vs.\ CMA$_{\text{non-conf}}$; Bootstrap: 1000 resamples of the ratio.}
\label{tab:cma-tests}
\centering
\scriptsize
\begin{tabular*}{\textwidth}{@{\extracolsep{\fill}}ccccc@{}}
\toprule
Step & $\theta$ & Perm.\ $p$ & Wilcoxon $p$ & Ratio 95\% CI \\
\midrule
0 & 0.3 & 0.044         & $4.9{\times}10^{-18}$ & [1.39, 1.59] \\
1 & 0.3 & 0.043         & $1.1{\times}10^{-19}$ & [1.40, 1.60] \\
2 & 0.3 & ${<}0.003$    & $6.8{\times}10^{-55}$ & [2.76, 3.13] \\
3 & 0.3 & ${<}0.002$    & $8.9{\times}10^{-107}$ & [2.64, 2.80] \\
3 & 0.5 & ${<}0.002$    & $4.8{\times}10^{-95}$ & [2.67, 2.87] \\
3 & 0.7 & ${<}0.005$    & $1.4{\times}10^{-31}$ & [3.01, 3.55] \\
4 & 0.3 & ${<}0.002$    & $5.8{\times}10^{-106}$ & \textbf{[4.11, 4.38]} \\
4 & 0.5 & ${<}0.002$    & $1.5{\times}10^{-98}$ & [3.93, 4.25] \\
4 & 0.7 & ${<}0.003$    & $1.0{\times}10^{-67}$ & [3.81, 4.24] \\
\bottomrule
\end{tabular*}
\end{table}

Table~\ref{tab:cma-detail} reports CMA across steps and thresholds; Table~\ref{tab:cma-tests} lists the corresponding hypothesis tests. At step~0, all three tests indicate a significant but mild effect ($p = 0.044$, Wilcoxon $p < 10^{-17}$, CI $[1.39,\, 1.59]$). From step~2 onward, the CMA ratio is strongly above $1$, reaching $4.2\times$ at step~4 ($\theta{=}0.3$, CI $[4.11,\, 4.38]$). Higher thresholds ($\theta{=}0.5,\, 0.7$) yield similar ratios at steps~3--4 but with fewer conflict edges, confirming robustness.

The shuffled baseline (Table~\ref{tab:cma-detail}, last column) closely tracks non-conflict CMA, confirming that the elevated CMA on conflict edges reflects genuine constraint structure rather than random variation in $\boldsymbol{\delta}$ magnitudes.

\subsection{Directional rather than Temperature Scaling}
\label{app:directional-vs-temperature}

\paragraph{Purpose.}
This experiment rules out the hypothesis that PISG's effect can be replicated by tuning the softmax temperature---a standard technique for adjusting distribution sharpness.

\paragraph{Definition.}
Temperature scaling adjusts the sharpness of a probability distribution by exponentiating and renormalizing:
\begin{equation}
p_e^{(T)} = \frac{p_e^{\,1/T}}{\sum_{e'} p_{e'}^{\,1/T}},
\end{equation}
where \(T < 1\) sharpens the distribution, \(T > 1\) smooths it, and \(T = 1\) is the identity. This is equivalent to dividing pre-softmax logits by \(T\).

\paragraph{Procedure.}
We sweep \(T \in \{0.3, 0.5, 0.8, 1.0, 1.5, 2.0\}\) on FastT2T for TSP-100 and MIS-ER (5-seed mean, seeds 41--45) and compare against PISG under default configuration.

\paragraph{Key argument.}
Temperature scaling is a rank-preserving scalar transformation: it cannot introduce instance-specific, edge-dependent directional corrections. The constraint-aligned signal characterized in \S\ref{app:tsp-signal-detail} (rank-dependent sign reversal captured by sc@2) and \S\ref{app:mis-signal-detail} (conflict-specific magnitude asymmetry captured by CMA) cannot arise from any scalar \(T\). Rank-dependent sign flips require non-monotonic transformations, and conflict-specific magnitude inflation requires per-element factors; neither is achievable by a single temperature parameter.

\paragraph{Results.}
No temperature improves either task, and PISG surpasses every temperature setting (TSP-100 gap: \(0.051\%\) vs.\ best-\(T\) \(0.079\%\); MIS-ER cost: \(38.89\) vs.\ best-\(T\) \(37.65\), i.e.\ gap $3.2\%$ below PISG).

\subsection{Mechanism Consistency Across Solvers}
\label{app:mechanism-cross-method}

\paragraph{Purpose.}
The signal characterization experiments in \S\ref{app:tsp-signal-detail}--\ref{app:mis-signal-detail} use FastT2T (\(T_s{=}5\), consistency model) as the primary analysis vehicle. To verify that the constraint-aligned signatures are not solver-specific, we repeat the full sc@2 and CMA analyses on StruDiCO (\(T_s{=}5\), variable-absorption diffusion with neighbor-aware decoding), which uses substantially different PISG configurations: adaptive perturbation with \(s{=}0.1\), \(L{=}7\), linear schedule for TSP; uniform perturbation with \(s{=}0.1\), \(L{=}9\), constant schedule for MIS-ER. All results use seed~42.

\paragraph{TSP: sc@2 on StruDiCO.}

\begin{table}[t]
\caption{StruDiCO sc@2 across denoising steps (TSP-100, 1280 instances, seed~42). Step indices start from~1 due to solver-specific trajectory conventions. Bootstrap 95\% CI over instances.}
\label{tab:strudico-sc2}
\centering
\scriptsize
\begin{tabular*}{\textwidth}{@{\extracolsep{\fill}}cccccc@{}}
\toprule
Step & \(t\) & sc@2 & Shuffled & Ratio & sc@2 95\% CI \\
\midrule
1 & 691 & \textbf{0.568} & 0.044 & \(\mathbf{12.9\times}\) & [0.559, 0.578] \\
2 & 413 & 0.277 & 0.045 & \(6.2\times\) & [0.269, 0.285] \\
3 & 191 & 0.021 & 0.024 & \(0.9\times\) & [0.019, 0.023] \\
4 & 49  & 0.003 & 0.016 & \(0.2\times\) & [0.002, 0.003] \\
\bottomrule
\end{tabular*}
\end{table}

Table~\ref{tab:strudico-sc2} shows that StruDiCO reproduces the same qualitative degree-2 sign pattern as FastT2T: sc@2 is highly significant at early steps (step~1: \(0.568\), \(12.9\times\) shuffle, permutation \(p < 0.01\); step~2: \(0.277\), \(6.2\times\), \(p < 0.01\)). The per-rank sign decomposition preserves the same structure: at step~1, rank~0 positive rate is \(79.0\%\), rank~1 is \(81.5\%\), and rank~2 negative rate is \(73.4\%\), all well above chance. The signal decays more rapidly than in FastT2T: by step~3, sc@2 falls below the shuffled baseline. This faster decay is consistent with StruDiCO's lower guidance scale (\(s{=}0.1\) vs.\ FastT2T's \(s{=}2.0\)) and its variable-absorption noise schedule.

\paragraph{MIS: CMA on StruDiCO.}

\begin{table}[t]
\caption{StruDiCO CMA ratio across denoising steps (MIS-ER, \(\theta{=}0.3\), 128 instances, seed~42). Bootstrap 95\% CI over instances.}
\label{tab:strudico-cma}
\centering
\scriptsize
\begin{tabular*}{\textwidth}{@{\extracolsep{\fill}}ccccccc@{}}
\toprule
Step & \(n_{\text{conflict}}\) & CMA\(_{\text{conf}}\) & CMA\(_{\text{non-conf}}\) & Ratio & Perm.\ \(p\) & 95\% CI \\
\midrule
0 &  11 & 0.061 & 0.021 & \(2.9\times\) & \(0.09\) & [2.01, 4.05] \\
1 &   8 & 0.078 & 0.024 & \(3.3\times\) & \(0.12\) & [2.04, 4.72] \\
2 &  28 & 0.103 & 0.029 & \(3.6\times\) & \({<}0.036\) & [2.60, 4.63] \\
3 &  70 & 0.161 & 0.031 & \(5.2\times\) & \({<}0.014\) & [4.62, 5.74] \\
4 & 105 & 0.140 & 0.022 & \(\mathbf{6.5\times}\) & \({<}0.010\) & \textbf{[5.79, 7.40]} \\
\bottomrule
\end{tabular*}
\end{table}

Table~\ref{tab:strudico-cma} shows that the CMA ratio grows monotonically from \(2.9\times\) at step~0 to \(6.5\times\) at step~4 (\(\theta{=}0.3\), CI \([5.79,\, 7.40]\)), exceeding FastT2T's peak of \(4.2\times\) at the same step. At higher thresholds, the pattern narrows: \(\theta{=}0.5\) reaches \(4.7\times\) (\(p < 0.015\)) at step~4, but \(\theta{=}0.7\) is not significant (ratio \(\approx 1.0\)), whereas FastT2T maintains \(4.0\times\) at \(\theta{=}0.7\). This indicates that StruDiCO concentrates its conflict-discrimination signal in the moderate-confidence boundary region rather than distributing it across all confidence levels.

\paragraph{Cross-solver summary.}
Both the degree-2 sign structure (TSP) and conflict-edge magnitude asymmetry (MIS) are confirmed on StruDiCO with full statistical tests, supporting that these constraint-aligned signatures arise from the task structure rather than from solver-specific artifacts. Appendix~\ref{app:gensco} reports the same two signatures on GenSCO, indicating that they are not unique to edge gating. The quantitative differences (faster sc@2 decay on TSP, stronger but narrower CMA on MIS) are consistent with the substantially different guidance scales and perturbation strategies used across the two solvers.

\subsection{Trajectory-Level Evidence}
\label{app:stepwise-decode}

This subsection provides two trajectory-level probes that together support a schedule-sensitive trajectory interpretation in Section~\ref{sec:signal-characterization}: stage-wise activation ablation compares early, late, and full-trajectory guidance, and stepwise greedy decoding shows guided intermediates can be individually worse while the final endpoint improves.

\paragraph{Stage-wise activation ablation.}

In the schedule-ablation protocol, \emph{Early} activates PISG on steps~\(\{0,1,2\}\), \emph{Late} activates steps~\(\{2,3,4\}\), \emph{All steps} activates the full trajectory, and \emph{Baseline} disables PISG.

\begin{table*}[t]
\caption{Stage-wise activation ablation under the schedule-ablation protocol (FastT2T, seed~42). Lower is better for TSP gap; higher is better for MIS solved cost.}
\label{tab:stepwise}
\centering
\scriptsize
\begin{tabular*}{\textwidth}{@{\extracolsep{\fill}}lcccc@{}}
\toprule
Task & Baseline & Early & Late & All steps \\
\midrule
TSP-100 gap\(\downarrow\) & 0.0860 & 0.0640 & 0.0643 & 0.0560 \\
MIS-ER cost\(\uparrow\) & 37.81 & 39.17 & 38.80 & 39.30 \\
\bottomrule
\end{tabular*}
\end{table*}

Together with the signal characterization (Section~\ref{sec:signal-characterization}) and the step-wise decode analysis below (Appendix~\ref{app:stepwise-decode}), these data support a trajectory-level interpretation: the benefit is schedule-sensitive and not explained by uniformly better intermediate greedy decodes.

\paragraph{Stepwise decode analysis.}
To test whether PISG uniformly improves the decoded solution at every intermediate step, we greedily decode \(\mathbf{x}_0\) predictions from all three branches (guided, normal, perturbed) at each of the \(T_s{=}5\) denoising steps, using matched \(\mathbf{x}_t\) inputs. All results are from FastT2T with seed~42.

\begin{table}[t]
\caption{TSP-100 stepwise decode: solution quality (gap\%, \(\downarrow\)) and structural feasibility (degree violation, \(\downarrow\)) at each step. Win\%: percentage of instances where guided gap is lower than normal gap. Steps~1--3: guided is worse in both metrics; only at step~4 does guided narrowly improve.}
\label{tab:stepwise-decode-tsp}
\centering
\scriptsize
\begin{tabular*}{\textwidth}{@{\extracolsep{\fill}}ccccccc@{}}
\toprule
& \multicolumn{3}{c}{Gap\% (\(\downarrow\))} & \multicolumn{2}{c}{Degree violation (\(\downarrow\))} & \\
\cmidrule(lr){2-4}\cmidrule(lr){5-6}
Step & Guided & Normal & \(\Delta\)(G\(-\)N) & Guided & Normal & Win\% \\
\midrule
0 & 2.023 & 2.023 & 0.000 & 0.0378 & 0.0378 & --- \\
1 & 2.170 & 1.982 & \(+\)0.188 & 0.0514 & 0.0378 & 24\% \\
2 & 1.994 & 1.731 & \(+\)0.263 & 0.0808 & 0.0307 & 22\% \\
3 & 0.696 & 0.448 & \(+\)0.249 & 0.1551 & 0.0083 & 9\% \\
4 & 0.176 & 0.206 & \(-\)0.030 & 0.0476 & 0.0017 & 3\% \\
\bottomrule
\end{tabular*}
\end{table}

\begin{table}[t]
\caption{MIS-ER stepwise decode: independent set size (\(\uparrow\)), conflict rate (\(\downarrow\)), and greedy skip ratio (\(\downarrow\)) at each step. Win\%: percentage of instances where guided IS size exceeds normal IS size. Guided achieves larger sets but with substantially worse feasibility.}
\label{tab:stepwise-decode-mis}
\centering
\scriptsize
\begin{tabular*}{\textwidth}{@{\extracolsep{\fill}}ccccccccc@{}}
\toprule
& \multicolumn{3}{c}{IS size (\(\uparrow\))} & \multicolumn{2}{c}{Conflict (\(\downarrow\))} & \multicolumn{2}{c}{Skip ratio (\(\downarrow\))} & \\
\cmidrule(lr){2-4}\cmidrule(lr){5-6}\cmidrule(lr){7-8}
Step & Guided & Normal & \(\Delta\) & Guided & Normal & Guided & Normal & Win\% \\
\midrule
0 & 24.03 & 1.78 & \(+\)22.25 & 0.0197 & 0.0048 & 0.286 & 0.001 & 98\% \\
1 & 23.99 & 1.88 & \(+\)22.11 & 0.0182 & 0.0047 & 0.262 & 0.001 & 98\% \\
2 & 19.95 & 4.71 & \(+\)15.24 & 0.0049 & 0.0032 & 0.088 & 0.000 & 92\% \\
3 & 36.54 & 31.99 & \(+\)4.55 & 0.0349 & 0.0028 & 0.691 & 0.156 & 67\% \\
4 & 38.35 & 37.80 & \(+\)0.55 & 0.0165 & 0.0017 & 0.514 & 0.167 & 40\% \\
\bottomrule
\end{tabular*}
\end{table}

\paragraph{TSP-100.}
Table~\ref{tab:stepwise-decode-tsp} shows that at step~0 guided \(\equiv\) normal (the linear schedule sets scale\(=0\)). At steps~1--3, guided intermediate solutions are \emph{worse} than normal: gap increases by \(+0.19\)--\(+0.26\) points, and degree violations are \(1.4\)--\(19\times\) higher (\(28\times\) at step~4). Only at the final step~4 does guided marginally improve (\(-0.03\) gap), though win rate is only \(3\%\). The consistently higher degree violations confirm that PISG's extrapolation produces intermediate predictions that are less structurally feasible when decoded greedily.

\paragraph{MIS-ER.}
Table~\ref{tab:stepwise-decode-mis} reveals a quality--feasibility trade-off. Guided predictions yield larger independent sets at every step, but with \(1.5\)--\(12\times\) more conflicts and higher skip rates during greedy decoding (\(3\)--\(4\times\) at steps~3--4; over \(380\times\) at steps~0--2, where the normal branch almost never skips, \(<0.001\)). The guided branch aggressively pushes probabilities above the selection threshold, inflating raw set size at the cost of feasibility.

\paragraph{Implications.}
Both tasks confirm that the final benefit of PISG does not arise from per-step decoded-solution improvement. On TSP, guided intermediate decodes are strictly worse; on MIS, the improvement in raw set size comes at the cost of much worse feasibility. The trajectory-level mechanism is the parsimonious explanation: PISG's per-step directional corrections reshape the denoising path in a way that benefits the final decode, even though intermediate decodes may appear worse in isolation.

\paragraph{Relation to the signal analysis.}
Section~\ref{sec:signal-characterization} and this subsection address different levels of the same question. Section~\ref{sec:signal-characterization} characterizes what the guidance direction encodes: \(\boldsymbol{\delta}_t\) concentrates at constraint boundaries rather than rescaling predictions isotropically. The stepwise decode tests a stronger hypothesis, namely that applying this direction improves the feasibility of each intermediate decode, and finds that it does not. The guidance direction is therefore focused on constraint-relevant positions, while step-wise feasibility improvement is not the route by which PISG improves the endpoint. The stage-wise ablation (Table~\ref{tab:stepwise}) is consistent with a trajectory-level effect: early-only and late-only guidance each improve over the baseline, and all-step guidance performs best. At the endpoint, the matched-noise probe (Table~\ref{tab:e22-tsp}) shows fewer decoder skips with guidance, for example \(43.3\to30.9\) on TSP-100.

\subsection{Association between Diagnostics and Final Quality}
\label{app:diagnostics-quality}

\paragraph{Purpose.}
sc@2 and CMA characterize the guidance direction. This subsection tests whether they also carry information about the final solution quality of PISG beyond instance difficulty. Easier instances tend to produce both stronger diagnostics and better solutions, so a raw correlation can reflect shared difficulty rather than an effect of the diagnosed signal.

\paragraph{Procedure.}
On FastT2T (seeds 41--45, aggregated per instance), we compute the Spearman correlation between each diagnostic and the final PISG quality, and the partial Spearman correlation that controls for the baseline's final quality on the same instance. TSP uses 1{,}280 TSP-100 instances and sc@2 at steps 0--2. MIS uses 127 MIS-ER instances and CMA at steps 3--4, the steps at which CMA is defined for nearly all instances. Intervals are 95\% bootstrap intervals over instances.

\begin{table}[ht]
\caption{Association between diagnostics and final PISG quality (FastT2T, seeds 41--45). Partial correlations control for the baseline's final quality on the same instance.}
\label{tab:diag-quality}
\centering\scriptsize
\begin{tabular*}{\linewidth}{@{\extracolsep{\fill}}lccc@{}}
\toprule
Diagnostic & Raw \(\rho\) & Partial \(\rho\) (controlling baseline) & 95\% CI of partial \(\rho\) \\
\midrule
TSP sc@2, steps 0--2 (\(n{=}1{,}280\)) & 0.44--0.45 & 0.03--0.05 & includes 0 \\
MIS CMA, steps 3--4 (\(n{=}127\)) & 0.34--0.39 & 0.33--0.36 & excludes 0 \\
\bottomrule
\end{tabular*}
\end{table}

\paragraph{Results.}
Table~\ref{tab:diag-quality} shows that the relationship depends on the task. On MIS, CMA keeps an association with the final independent-set size after controlling for baseline quality (partial \(\rho=0.33\)--\(0.36\), \(p<0.001\)): among instances where the baseline reaches similar set sizes, those with stronger conflict-edge asymmetry still tend to produce larger guided sets. CMA is therefore an outcome-linked diagnostic, but this association does not establish it as a causal mediator. On TSP, the raw correlation of sc@2 with final quality (\(\rho\approx0.45\)) nearly vanishes once baseline quality is controlled (partial \(\rho\approx0.03\)--\(0.05\)). PISG and baseline quality track each other closely across instances (\(\rho=0.93\)), so the raw association reflects shared instance difficulty. sc@2 remains a structural diagnostic of the guidance direction but does not predict per-instance benefit.

\subsection{No Subtour-Level Signal in TSP}
\label{app:subtour}

\paragraph{Purpose.}
The TSP diagnostic in Section~\ref{sec:signal:tsp} measures the local degree-2 boundary. This subsection tests whether \(\boldsymbol{\delta}_t\) also carries subtour-level information, that is, whether guided predictions help the greedy decoder avoid closing cycles before all nodes are connected.

\paragraph{Procedure.}
During greedy decoding, a candidate edge is rejected as fragment-closing when it would close a cycle among a proper subset of the nodes. If \(\boldsymbol{\delta}_t\) carried subtour information, decoding the guided prediction should produce fewer such rejections \(R\) than decoding the normal prediction. We decode the normal, guided, and shuffled-\(\boldsymbol{\delta}_t\) predictions at each step on FastT2T TSP-100 (seeds 41--45, 1{,}280 instances per seed, ten within-instance shuffles) and report \(\Delta R_t = R^{\text{normal}}_t - R^{\text{guided}}_t\) (\(>0\): fewer rejections with guidance) with 95\% hierarchical bootstrap intervals. Step~0 is omitted because the three predictions coincide there.

\begin{table}[ht]
\caption{Fragment-closing rejections per instance during greedy decoding (FastT2T TSP-100, seeds 41--45). \(\Delta R_t>0\) would indicate that guidance reduces premature cycle closure.}
\label{tab:subtour}
\centering\scriptsize
\begin{tabular*}{\linewidth}{@{\extracolsep{\fill}}ccccc@{}}
\toprule
Step & Normal \(R\) & Guided \(R\) & Shuffled \(R\) & \(\Delta R_t\) (95\% CI) \\
\midrule
1 & 0.5670 & 0.5345 & 0.5623 & \(+\)0.033 [\({-}\)0.007, 0.075] \\
2 & 0.4169 & 0.4406 & 0.4116 & \({-}\)0.024 [\({-}\)0.060, 0.015] \\
3 & 0.0973 & 0.1259 & 0.1022 & \({-}\)0.029 [\({-}\)0.046, \({-}\)0.012] \\
4 & 0.0325 & 0.0311 & 0.0335 & \(+\)0.001 [\({-}\)0.007, 0.011] \\
\bottomrule
\end{tabular*}
\end{table}

\paragraph{Results.}
Table~\ref{tab:subtour} provides no evidence of subtour-level signal. Three of the four intervals include zero, and at step~3 guidance increases fragment-closing rejections. The TSP mechanism evidence in this paper therefore concerns local degree-2 boundary sensitivity, not subtour awareness.

%% file: neurips_2026.bbl
\begin{thebibliography}{25}
\providecommand{\natexlab}[1]{#1}
\providecommand{\url}[1]{\texttt{#1}}
\expandafter\ifx\csname urlstyle\endcsname\relax
  \providecommand{\doi}[1]{doi: #1}\else
  \providecommand{\doi}{doi: \begingroup \urlstyle{rm}\Url}\fi

\bibitem[Ahn et~al.(2024)Ahn, Cho, Min, Jang, Kim, Kim, Park, Jin, and
  Kim]{ahn2024pag}
D.~Ahn, H.~Cho, J.~Min, W.~Jang, J.~Kim, S.~Kim, H.~H. Park, K.~H. Jin, and
  S.~Kim.
\newblock Self-rectifying diffusion sampling with perturbed-attention guidance.
\newblock In \emph{European Conference on Computer Vision}, 2024.

\bibitem[Ahn et~al.(2025)Ahn, Kang, Lee, Kim, Jang, Min, Lee, Paul, and
  Kim]{ahn2025headhunter}
D.~Ahn, J.~Kang, S.~Lee, M.~Kim, W.~Jang, J.~Min, S.~Lee, S.~Paul, and S.~Kim.
\newblock Where and how to perturb: On the design of perturbation guidance in
  diffusion and flow models.
\newblock In \emph{Advances in Neural Information Processing Systems}, 2025.

\bibitem[Applegate et~al.(2006)Applegate, Bixby, Chv{\'a}tal, and
  Cook]{applegate2006concorde}
D.~Applegate, R.~E. Bixby, V.~Chv{\'a}tal, and W.~J. Cook.
\newblock Concorde {TSP} solver, 2006.
\newblock URL \url{https://www.math.uwaterloo.ca/tsp/concorde.html}.
\newblock Software package.

\bibitem[Bansal et~al.(2024)Bansal, Chu, Schwarzschild, Sengupta, Goldblum,
  Geiping, and Goldstein]{bansal2023universal}
A.~Bansal, H.-M. Chu, A.~Schwarzschild, R.~Sengupta, M.~Goldblum, J.~Geiping,
  and T.~Goldstein.
\newblock Universal guidance for diffusion models.
\newblock In \emph{International Conference on Learning Representations}, 2024.
\newblock Earlier versions appeared in CVPR Workshops 2023 and
  arXiv:2302.07121.

\bibitem[Bresson and Laurent(2017)]{bresson2017residual}
X.~Bresson and T.~Laurent.
\newblock Residual gated graph convnets.
\newblock \emph{arXiv preprint arXiv:1711.07553}, 2017.

\bibitem[Chen et~al.(2026)Chen, Li, and Yan]{chen2026maskco}
L.~Chen, Y.~Li, and J.~Yan.
\newblock {MaskCO}: Masked generation drives effective representation learning
  and exploiting for combinatorial optimization.
\newblock In \emph{International Conference on Learning Representations}, 2026.

\bibitem[Dhariwal and Nichol(2021)]{dhariwal2021diffusion}
P.~Dhariwal and A.~Q. Nichol.
\newblock Diffusion models beat {GANs} on image synthesis.
\newblock In \emph{Advances in Neural Information Processing Systems}, 2021.

\bibitem[Han et~al.(2026)Han, Zhang, and Wang]{han2026cdg}
S.~Han, Y.~Zhang, and H.~Wang.
\newblock Guiding diffusion models with semantically degraded conditions.
\newblock In \emph{IEEE/CVF Conference on Computer Vision and Pattern
  Recognition}, 2026.

\bibitem[He et~al.(2026)He, Rojas, and Tao]{he2026guidance}
Y.~He, K.~Rojas, and M.~Tao.
\newblock What exactly does guidance do in masked discrete diffusion models.
\newblock In \emph{International Conference on Learning Representations}, 2026.

\bibitem[Ho and Salimans(2022)]{ho2022cfg}
J.~Ho and T.~Salimans.
\newblock Classifier-free diffusion guidance.
\newblock \emph{arXiv preprint arXiv:2207.12598}, 2022.
\newblock A short version appeared in the NeurIPS 2021 Workshop on Deep
  Generative Models and Downstream Applications.

\bibitem[Hong(2024)]{hong2024smoothed}
S.~Hong.
\newblock Smoothed energy guidance: Guiding diffusion models with reduced
  energy curvature of attention.
\newblock In \emph{Advances in Neural Information Processing Systems}, 2024.

\bibitem[Hong et~al.(2023)Hong, Lee, Jang, and Kim]{hong2023improving}
S.~Hong, G.~Lee, W.~Jang, and S.~Kim.
\newblock Improving sample quality of diffusion models using self-attention
  guidance.
\newblock In \emph{International Conference on Computer Vision}, 2023.

\bibitem[Lei et~al.(2026)Lei, Zhou, Li, Chen, and Farnia]{lei2025difuada}
H.~Lei, K.~Zhou, Y.~Li, Z.~Chen, and F.~Farnia.
\newblock Boosting cross-problem generalization in diffusion-based neural
  combinatorial solver via inference time adaptation.
\newblock \emph{Proceedings of the AAAI Conference on Artificial Intelligence},
  40\penalty0 (43):\penalty0 36289--36297, 2026.
\newblock \doi{10.1609/aaai.v40i43.40948}.

\bibitem[Li et~al.(2026)Li, Luo, Chen, Ma, and Qi]{chung2024selfguidance}
T.~Li, W.~Luo, Z.~Chen, L.~Ma, and G.-J. Qi.
\newblock Self-guidance: Boosting flow and diffusion generation on their own.
\newblock \emph{IEEE Transactions on Pattern Analysis and Machine
  Intelligence}, 48\penalty0 (1):\penalty0 781--791, 2026.
\newblock \doi{10.1109/TPAMI.2025.3611831}.

\bibitem[Li et~al.(2023)Li, Guo, Wang, and Yan]{li2023t2t}
Y.~Li, J.~Guo, R.~Wang, and J.~Yan.
\newblock {T2T}: From distribution learning in training to gradient search in
  testing for combinatorial optimization.
\newblock In \emph{Advances in Neural Information Processing Systems}, 2023.

\bibitem[Li et~al.(2024)Li, Guo, Wang, Zha, and Yan]{li2024fastt2t}
Y.~Li, J.~Guo, R.~Wang, H.~Zha, and J.~Yan.
\newblock Fast {T2T}: Optimization consistency speeds up diffusion-based
  training-to-testing solving for combinatorial optimization.
\newblock In \emph{Advances in Neural Information Processing Systems}, 2024.

\bibitem[Li et~al.(2025)Li, Chen, Wang, Wang, and Yan]{li2025gensco}
Y.~Li, L.~Chen, H.~Wang, R.~Wang, and J.~Yan.
\newblock Generation as search operator for test-time scaling of
  diffusion-based combinatorial optimization.
\newblock In \emph{Advances in Neural Information Processing Systems}, 2025.

\bibitem[Ma et~al.(2025)Ma, Pan, Li, and Yan]{ma2025coexpander}
J.~Ma, W.~Pan, Y.~Li, and J.~Yan.
\newblock {COExpander}: Adaptive solution expansion for combinatorial
  optimization.
\newblock In \emph{International Conference on Machine Learning}, 2025.

\bibitem[Qiu et~al.(2022)Qiu, Sun, and Yang]{qiu2022dimes}
R.~Qiu, Z.~Sun, and Y.~Yang.
\newblock {DIMES}: A differentiable meta solver for combinatorial optimization
  problems.
\newblock In \emph{Advances in Neural Information Processing Systems}, 2022.

\bibitem[Reinelt(1991)]{reinelt1991tsplib}
G.~Reinelt.
\newblock {TSPLIB}---a traveling salesman problem library.
\newblock \emph{ORSA Journal on Computing}, 3\penalty0 (4):\penalty0 376--384,
  1991.

\bibitem[Schiff et~al.(2025)Schiff, Sahoo, Phung, Wang, Boshar, Dalla-Torre,
  de~Almeida, Rush, Pierrot, and Kuleshov]{schiff2025simple}
Y.~Schiff, S.~S. Sahoo, H.~Phung, G.~Wang, S.~Boshar, H.~Dalla-Torre, B.~P.
  de~Almeida, A.~Rush, T.~Pierrot, and V.~Kuleshov.
\newblock Simple guidance mechanisms for discrete diffusion models.
\newblock In \emph{International Conference on Learning Representations}, 2025.

\bibitem[Sun and Yang(2023)]{sun2023difusco}
Z.~Sun and Y.~Yang.
\newblock {DIFUSCO}: Graph-based diffusion solvers for combinatorial
  optimization.
\newblock In \emph{Advances in Neural Information Processing Systems}, 2023.

\bibitem[Wang et~al.(2025)Wang, Li, Yan, and Chang]{wang2025strudico}
Y.~Wang, Y.~Li, J.~Yan, and Y.~Chang.
\newblock {StruDiCO}: Structured denoising diffusion with gradient-free
  inference-stage boosting for memory and time efficient combinatorial
  optimization.
\newblock In \emph{Advances in Neural Information Processing Systems}, 2025.

\bibitem[Wang et~al.(2026)Wang, Li, Ma, Yan, and Chang]{wang2026nexco}
Y.~Wang, Y.~Li, J.~Ma, J.~Yan, and Y.~Chang.
\newblock Native adaptive solution expansion for diffusion-based combinatorial
  optimization.
\newblock In \emph{International Conference on Learning Representations}, 2026.

\bibitem[Ye et~al.(2024)Ye, Lin, Han, Xu, Liu, Liang, Ma, Zou, and
  Ermon]{ye2024tfg}
H.~Ye, H.~Lin, J.~Han, M.~Xu, S.~Liu, Y.~Liang, J.~Ma, J.~Zou, and S.~Ermon.
\newblock {TFG}: Unified training-free guidance for diffusion models.
\newblock In \emph{Advances in Neural Information Processing Systems}, 2024.
\newblock Spotlight.

\end{thebibliography}
